\pdfoutput=1

\documentclass[11pt]{article}

\usepackage[final]{acl}

\usepackage{times}
\usepackage{latexsym}
\usepackage{amsmath}
\usepackage{amssymb}
\usepackage{booktabs} 
\usepackage{enumitem}
\usepackage{graphicx}
\usepackage{pifont}
\usepackage{color}
\usepackage{titletoc}
\usepackage{colortbl}
\usepackage{wrapfig}  
\usepackage{lipsum} 
\usepackage{multirow}
\usepackage{listings}
\usepackage{array}
\usepackage{booktabs} 
\usepackage{caption}  
\usepackage{multicol}

\usepackage{bbm}

\usepackage{inconsolata}
\usepackage{ulem}
\usepackage{array}
\usepackage{amsfonts}
\usepackage{algorithm}
\usepackage{algpseudocode}
\usepackage{float}

\usepackage{caption}
\usepackage[T1]{fontenc}

\usepackage[utf8]{inputenc}

\usepackage{microtype}

\usepackage{inconsolata}

\usepackage{graphicx}

\definecolor{myGreen}{RGB}{34, 139, 34}
\definecolor{myRed}{HTML}{FF6347}

\newcommand{\cmark}{\textcolor{myGreen}{\ding{51}}}
\newcommand{\xmark}{\textcolor{myRed}{\ding{55}}}
\newcommand{\ours}{\textsc{MMRC}} 
\newcommand{\platform}{\textit{DialogFlow}} 
\newcommand{\strategy}{\textsc{Note-taking}}

\title{MMRC: A Large-Scale Benchmark for Understanding Multimodal Large Language Model in Real-World  Conversation}

\author{Haochen Xue$^{1,2*}$, Feilong Tang$^{1,3,4*}$, Ming Hu$^{1,3*}$,Yexin Liu$^{5}$,Qidong Huang$^{1,6}$, Yulong Li$^{1,2}$\\ \bf  Chengzhi Liu$^{2}$, Zhongxing Xu$^{3}$, Chong Zhang$^{2}$, Chun-Mei Feng$^{7}$, Yutong Xie$^{4}$\\ \bf  Imran Razzak$^{4}$, Zongyuan Ge$^{3\dag}$, Jionglong Su$^{2\dag}$, Junjun He$^{1\dag}$, Yu Qiao$^{1}$ \\ 
$^{1}$  Shanghai Artificial Intelligence Laboratory, $^{2}$ Xi'an Jiaotong-Liverpool University, \\ $^{3}$ Monash University, $^{4}$ MBZUAI, $^{5}$HKUST, $^{6}$USTC, $^{7}$IHPC, A*STAR}

\begin{document}
\maketitle

\def\thefootnote{$^{*}$}\footnotetext{Equal contribution. {$\dag$} Corresponding authors. \\
\hspace*{1.8em}\url{https://github.com/tiuxuxsh76075/MMRC}}\def\thefootnote{\arabic{footnote}}

\begin{abstract}
Recent multimodal large language models (MLLMs) have demonstrated significant potential in open-ended conversation, generating more accurate and personalized responses. However, their abilities to memorize, recall, and reason in sustained interactions within real-world scenarios remain underexplored. This paper introduces \ours{}, a \textbf{M}ulti-\textbf{M}odal \textbf{R}eal-world \textbf{C}onversation benchmark for evaluating six core open-ended abilities of MLLMs: information extraction, multi-turn reasoning, information update, image management, memory recall, and answer refusal. With data collected from real-world scenarios, \ours{} comprises 5,120 conversations and 28,720 corresponding manually labeled questions, posing a significant challenge to existing MLLMs. Evaluations on 20 MLLMs in \ours{} indicate an accuracy drop during open-ended interactions. We identify four common failure patterns: long-term memory degradation, inadequacies in updating factual knowledge, accumulated assumption of error propagation, and reluctance to “say no.” To mitigate these issues, we propose a simple yet effective \strategy{} strategy, which can record key information from the conversation and remind the model during its responses, enhancing conversational capabilities. Experiments across six MLLMs demonstrate significant performance improvements. 
\end{abstract}

\vspace{-0.5em}
\section{Introduction}
\vspace{-0.5em}
Open-ended conversations (OEC) are the most common form of interaction between humans and Multimodal Large Language Models (MLLMs), representing a crucial feature of Artificial General Intelligence (AGI)~\citep{kilmllm,fei2024multimodal}. These conversations are entirely determined by the user's intention, rather than by system rules or predefined patterns~\citep{decker2022topic,zheng2023lmsys, liu2024convbench}. Furthermore, individual differences shape each conversation’s distinct linguistic style and user-specific preferences~\citep{chaves2022chatbots,ma2024goal,tan2025can}.

\begin{table*}[t]
    \centering
    \footnotesize
    \setlength{\tabcolsep}{3pt}
    \vspace{-1em}
    \resizebox{\linewidth}{!}{
    \begin{tabular}{l|cccccccc|c|c|c|c|c|c}
    \toprule
    \multirow{2}{*}{\textbf{Dataset}} & \multirow{2}{*}{\textbf{Dialog}} & \multirow{2}{*}{\textbf{A-Turns}} & \multirow{2}{*}{\textbf{Img}} & \multirow{2}{*}{\textbf{Multi-Img}} & \multirow{2}{*}{\textbf{Domains}} & \multirow{2}{*}{\textbf{Gen-Method}} & \multirow{2}{*}{\textbf{Temp-Type}} & \multicolumn{7}{c}{\textbf{ Core Conversation Abilities}}\\ 
    &  &  & &&&& &  & \textbf{IE} & \textbf{CR} & \textbf{IU}& \textbf{IM} & \textbf{MR} & \textbf{AR} \\\midrule
       MT-Bench~\cite{zheng2023judging} &80 & 2.0 &  \xmark & \xmark & 8 &  GPT-4 & Fixed &  & \cmark & \cmark & \xmark & \xmark & \xmark & \xmark \\
       MT-Bench-101~\citep{bai2024mt} & 1388 & 3.1 &  \xmark & \xmark & 13 & GPT-4o & Fixed &  & \cmark & \cmark & \cmark & \xmark & \xmark & \xmark  \\
       LongMemEval~\citep{wu2024longmemeval} & 500 & 6.1 &  \xmark & \xmark & 164 & GPT-4 & Fixed &  & \cmark & \cmark & \cmark & \xmark & \cmark & \cmark \\
       DialogBench~\citep{ou2023dialogbench} & 9811 & 7.6 &  \xmark & \xmark & 36  & GPT-4 & Fixed &  & \cmark & \cmark & \cmark & \xmark & \cmark & \xmark \\
       Farm~\citep{xu2023earth} & 1952 & 7.4 &  \xmark & \xmark & 5 & GPT-4o & Fixed &  & \xmark & \cmark & \xmark & \xmark & \xmark & \cmark \\
       MT-Bench ++~\citep{sun2024parrot} & - & 8 &  \cmark & \xmark & 13 & GPT-4o & Fixed &  & \cmark & \cmark & \cmark & \xmark & \cmark & \xmark \\
       MMR~\citep{liu2024seeing} & 84 & 2.3 & \cmark & \xmark & 12 & GPT-4 & Fixed &  & \cmark & \cmark & \xmark & \xmark & \xmark & \xmark \\
       EvalDial~\citep{park2024mitigating} & 500 & 2.7 & \cmark & \xmark & 96 & GPT-4 & Fixed &  & \cmark & \cmark & \xmark & \xmark & \xmark & \cmark \\
       ConvBench~\citep{liu2024convbench} & 577 & 3.0 & \cmark & \xmark & 197 & GPT-4o & Fixed &  & \cmark & \cmark & \xmark & \xmark & \xmark & \xmark \\
       MMDU Benchmark~\citep{liu2024mmdu} & 110 & 15 & \cmark & \cmark & 219 & GPT-4o & Fixed &  & \cmark & \cmark & \xmark & \cmark & \cmark & \xmark \\
\midrule
       \textbf{\ours{}\ (Ours)} & 5120 & \textbf{15.2} & \cmark & \cmark & \textbf{874} & \textbf{Real-conv} & \textbf{Open-ended} &  & \cmark& \cmark& \cmark& \cmark& \cmark & \cmark\\

    \bottomrule
    \end{tabular}
    }
    \vspace{-0.2cm}
    \caption{The comparison between \ours{}\ with other conversation benchmarks. 
    \textbf{Dialog}: the number of dialogs; \textbf{A-Turns}: average turns per conversation; \textbf{Img}: support for single image input; \textbf{Multi-Img}: support for multiple images input; \textbf{Domains}: the number of covered domains; \textbf{Gen-Method}: generation method; \textbf{Temp-Type}: dialogue template type; Finally, we the coverage of six core abilities: information extraction (IE), cross-turn reasoning (CR), information update (IU), image management (IM), memory recall (MR), answer refusal (AR).}
    \vspace{-0.4cm}
    \label{tab:data_compare}
\end{table*}

\begin{figure}[t]
\centering
\vspace{-0.4cm}
\includegraphics[width=\linewidth]{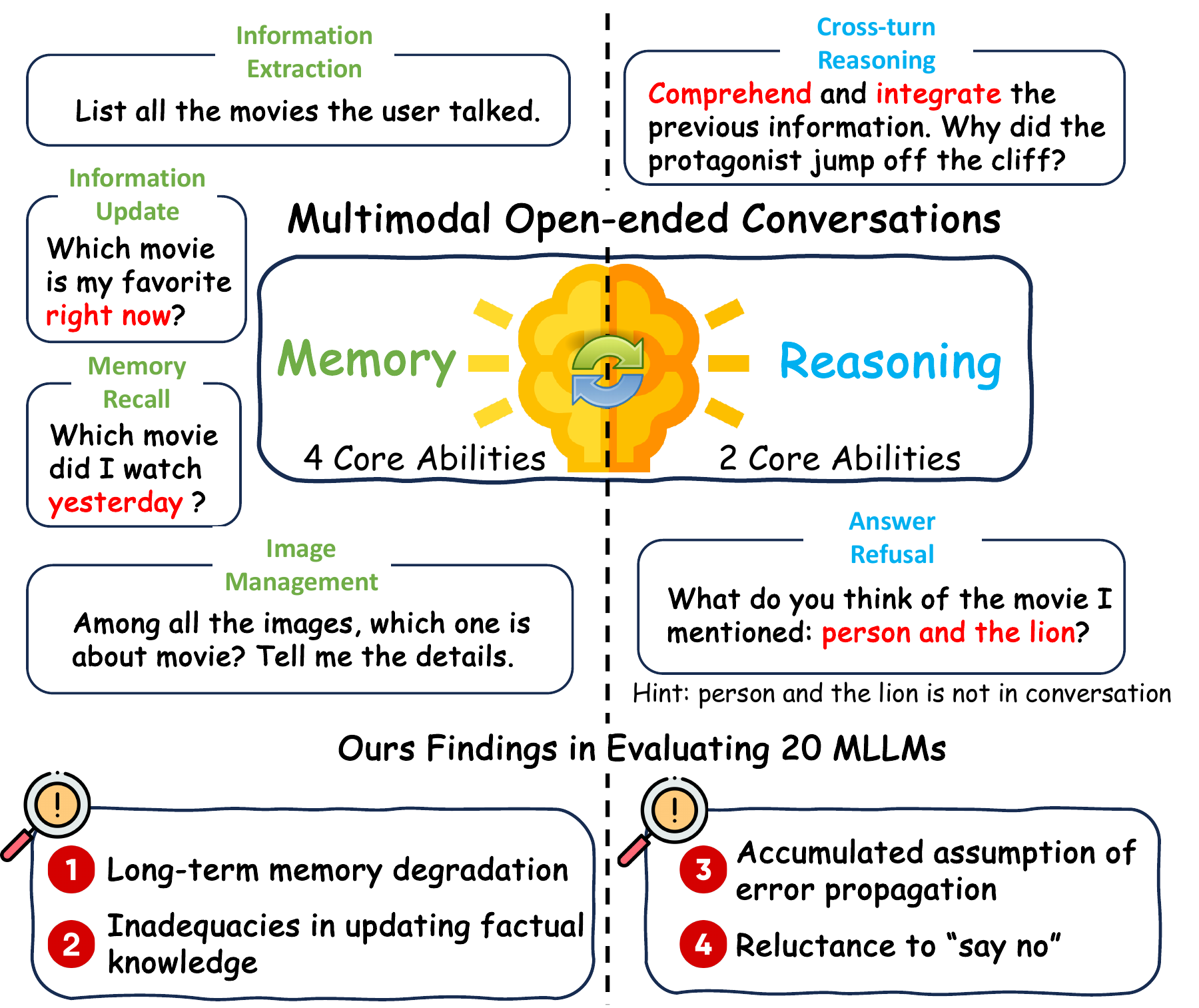}
\caption{Illustration of the six core multimodal open-ended conversation abilities in the \ours{} benchmark.}
\label{fig:fig1}
\vspace{-0.5cm}
\end{figure}

However, existing conversation benchmarks~\citep{bai2024mt,liu2024seeing,xu2023earth,liu2024convbench} fail to comprehensively evaluate MLLMs’ abilities to memorize, recall, and reason in sustained interactions in OEC. These benchmarks exhibit two primary limitations: \textit{(i):} their reliance on prompt templates for dataset generation limits the diversity of data domains and conversation lengths, leading to repetitive and overly structured dialogues that fail to reflect the complexities of long-term user-AI interactions. \textit{(ii):} they only cover a limited subset of the memory capabilities required to leverage dynamic, ever-changing, and accumulative information in long-term interactions, failing to evaluate the ability to recall multimodal information or reason with updated information.

Based on these limitations, we develop \platform, a free online dialogue platform featuring 20 cutting-edge MLLMs to collect diverse, real-world conversation data. Through \platform, we construct \ours{}, the first \textbf{M}ulti-\textbf{M}odal \textbf{R}eal-world \textbf{C}onversation benchmark, which includes 5,120 carefully selected dialogues. Drawing on cognitive studies of human conversation~\citep{clark2019makes,liddicoat2021introduction}, an evaluation framework consisting of 28,720 manually annotated questions is designed to assess MLLMs' six core abilities in OEC, as illustrated in Fig.~\ref{fig:fig1}: \textit{information extraction}, \textit{cross-turn reasoning}, \textit{information update}, \textit{image management}, \textit{long-term memory recall}, and \textit{answer refusal}. To achieve this, multiple evaluation metrics, including GPT-based evaluations, human evaluation, and objective precision metrics, are employed to ensure a comprehensive and objective assessment.

We extensively evaluate 20 mainstream MLLMs and observe that they, including advanced GPT-4o~\citep{islam2024gpt}, fail to consistently deliver accurate and reliable responses over extended interactions. Furthermore, certain open-source MLLMs demonstrate a limited capacity for long-term conversations in real scenarios. Through analyzing models' error answers, we identify four common failure patterns: (1) \uline{Long-term memory degradation}, where MLLMs' memory of facts from earlier conversations becomes vague, leading to inconsistent responses with prior turns. (2) \uline{Inadequacie in updating factual knowledge}, where MLLMs exhibit a failure to integrate new facts effectively, still continuing to rely on outdated information. (3) \uline{Accumulated assumption of error propagation}, where erroneous assumptions made while integrating information from earlier turns propagate into later turns, leading to an interrupted reasoning chain. (4) \uline{Reluctance to “say no”}, where MLLMs show an inability to decline to provide an answer in OEC when the context is insufficient.

To mitigate this, we propose a simple yet effective method called \strategy{}. This strategy systematically stores key user preferences and facts throughout the conversation. When the model is tasked with evaluation queries, the recorded information is transformed into structured prompts, providing supplementary context to improve the accuracy and coherence of the MLLMs' responses. Experiment results across six MLLMs demonstrated that this strategy significantly enhances the models' overall conversational capabilities.

In summary, our contributions are four-fold: (1) We introduce the first multi-modal open-ended conversation (OEC) benchmark \ours{}, providing a comprehensive evaluation of MLLMs' performance in practical settings. (2) We propose six core abilities of the model in OEC, covering broader aspects than existing benchmarks. (3) Using our evaluation framework, we analyze 20 state-of-the-art MLLMs and identify four failure patterns in OEC, providing insights to inspire future research. (4) We propose \strategy{}, which improves conversational capabilities by storing key user preferences and facts and using structured prompts to assist MLLMs in generating responses.

\vspace{-0.2cm}
\section{Related Work}
\vspace{-0.1cm}
\textbf{Multimodal Large Language Model.}
Building on large language models, multimodal large language models (MLLMs) have exhibited remarkable capabilities~\citep{kilmllm,cui2024survey,qin2025survey}, achieving state-of-the-art performance across various downstream tasks, including visual grounding~\citep{li2024groundinggpt,xu2024mc}, object detection~\citep{zang2024contextual,wu2025dettoolchain}, visual question answering (VQA)~\citep{kuang2024natural,xu2024mlevlm,demllms}, and instruction following~\citep{li2023fine,sun2024parrot,wei2024demonstrative}. Their outstanding performance underscores their pivotal role in AGI~\cite{zhang2024improving}.

\begin{figure*}[!t]
 \centering
  \captionsetup{justification=raggedright, singlelinecheck=false}
  \includegraphics[width=0.91\textwidth]{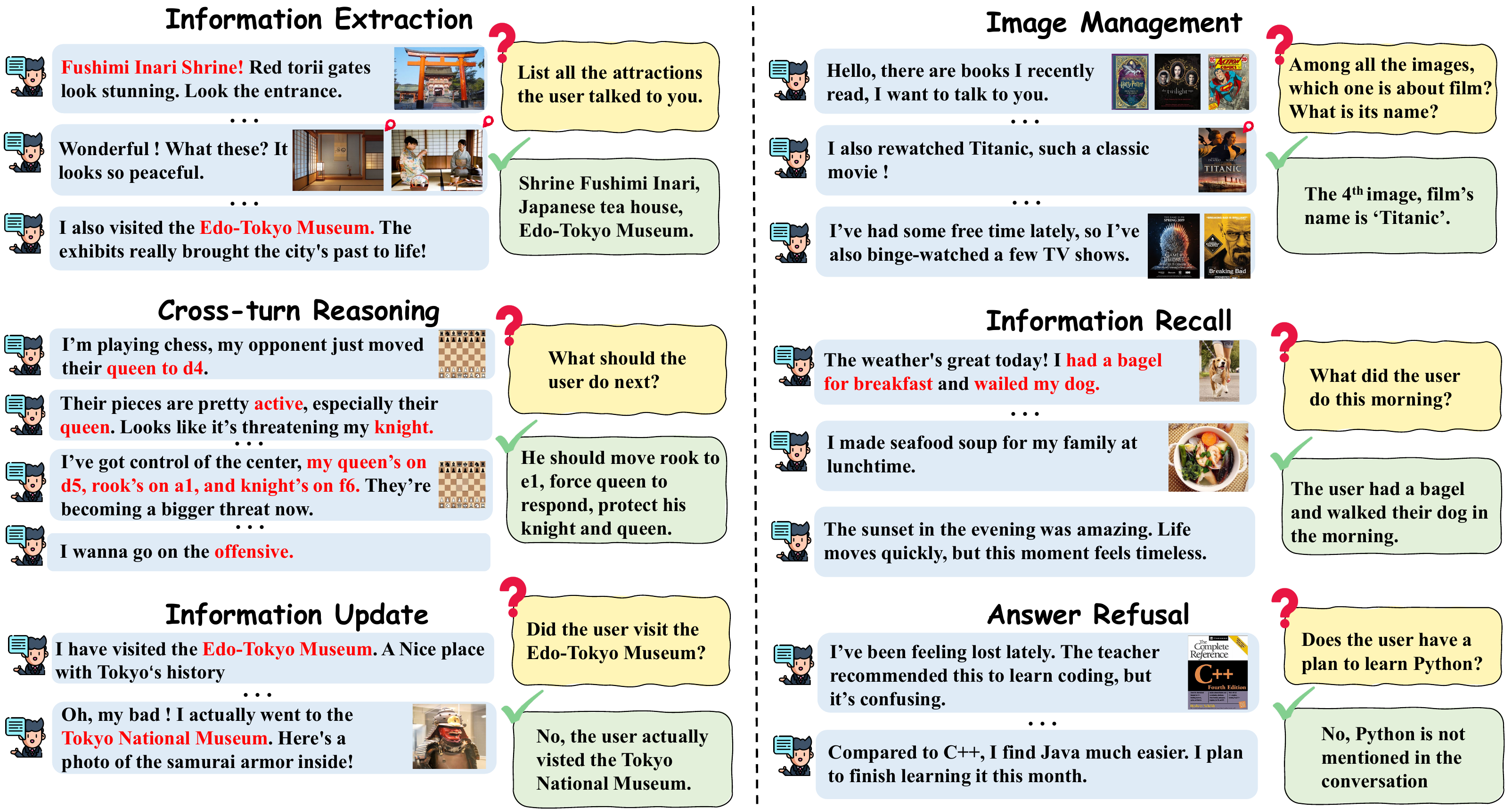} 
  \vspace{-0.2cm}
  \caption{A sample from the \ours{}, featuring a multi-turn open-ended conversation with six human-annotated questions and answers, designed to assess the ability of MLLMs in open-ended conversations.}
  \vspace{-0.4cm}
  \label{fig:6ability}
\end{figure*}

\noindent \textbf{Benchmarks for Long-Term Conversation.}
MT-Bench~\citep{zheng2023judging} is a pioneering two-turn dialogue dataset generated by GPT, covering eight domain tasks. MT-Bench-101~\citep{bai2024mt} and Bench++~\citep{sun2024parrot} expand the dataset size and add more domains, enhancing evaluation depth. In parallel, Farm~\cite{xu2023earth}, EvalDial~\citep{park2024mitigating}, and MMR~\cite{liu2024seeing} examine model robustness in multi-turn dialogue scenarios using fixed dialogue formats. ConvBench~\citep{liu2024convbench} evaluates models' perception, reasoning, and creation abilities through structured three-turn dialogues, exploring their interrelations. DialogBench~\citep{ou2023dialogbench} and LongMemEval~\citep{wu2024longmemeval} focus on evaluating models' abilities in context understanding and memory retention during GPT-generated dialogues. MMDU~\citep{liu2024mmdu} evaluates the understanding and instruction-following abilities in GPT-generated multi-image, multi-turn dialogues. Table~\ref{tab:data_compare} compares \ours{} with previous works, highlighting its advantages in: (1) naturally open dialogue format with longer and more diverse conversations. (2) holistically covering critical abilities in memorization, recall, and reasoning in a uniquely challenging way (further examples in Fig.~\ref{fig:6ability}).

\vspace{-0.2cm}
\section{The \ours{}\ }
\vspace{-0.1cm}

\subsection{Problem Formulation}\label{3.1}
The evaluation of \ours{} requires a triplet instance $(S, q, a)$, where $S$ represents the dialogue history, $q$ is a set of evaluation questions assessing specific conversational abilities, and $a$ is the ground truth answers. Specifically, $S = \{(t_i, R_i)\}_{i=1}^{n}$ denotes an $n$-turn dialogue history, where $t_i = (\text{text}_i, \text{image}_i)$ represents the user query with text, images, or both, and $R_i$ is the model's response at turn $i$. With our MMRC setup, given the dialogue context $S$, the model is tasked to answer a set of six evaluation questions, $q = \{q_i\}_{i=1}^{T}$, where $T = 6$, each designed to assess a specific ability. The model's responses, denoted as $p = \{p_i\}_{i=1}^{T}$, are then individually compared against human-annotated ground truth answers, $a = \{a_i\}_{i=1}^{T}$, to evaluate its performance in OEC.

We summarize the memorization, recall, and reasoning abilities required by MLLMs in OEC, as illustrated in Fig.~\ref{fig:6ability}, with details as follows.

\noindent\textbf{Information Extraction (IE)}: Ability to retrieve specific information from the conversation history, which includes both textual and visual content.

\noindent\textbf{Cross-turn Reasoning (CR)}: Ability to comprehend and integrate information across multiple dialogue turns to answer complex questions.

\noindent\textbf{Information Update (IU)}: Ability to track and update knowledge dynamically by recognizing changes in user information and factual updates.

\noindent\textbf{Image Management (IM)}: Ability to store and manage visual information by retaining specific image details and maintaining accurate attribution.

\noindent\textbf{Memory Recall (MR)}: Ability to maintain and retrieve memory of previous interactions throughout the conversation and recall user-specific details.

\noindent\textbf{Answer Refusal (AR)}: Ability to refrain from answering questions that involve unknown information, \textit{i.e.,} absent from the interaction history.

\begin{figure}[!t]  
  \centering
  \includegraphics[width=0.95\linewidth]{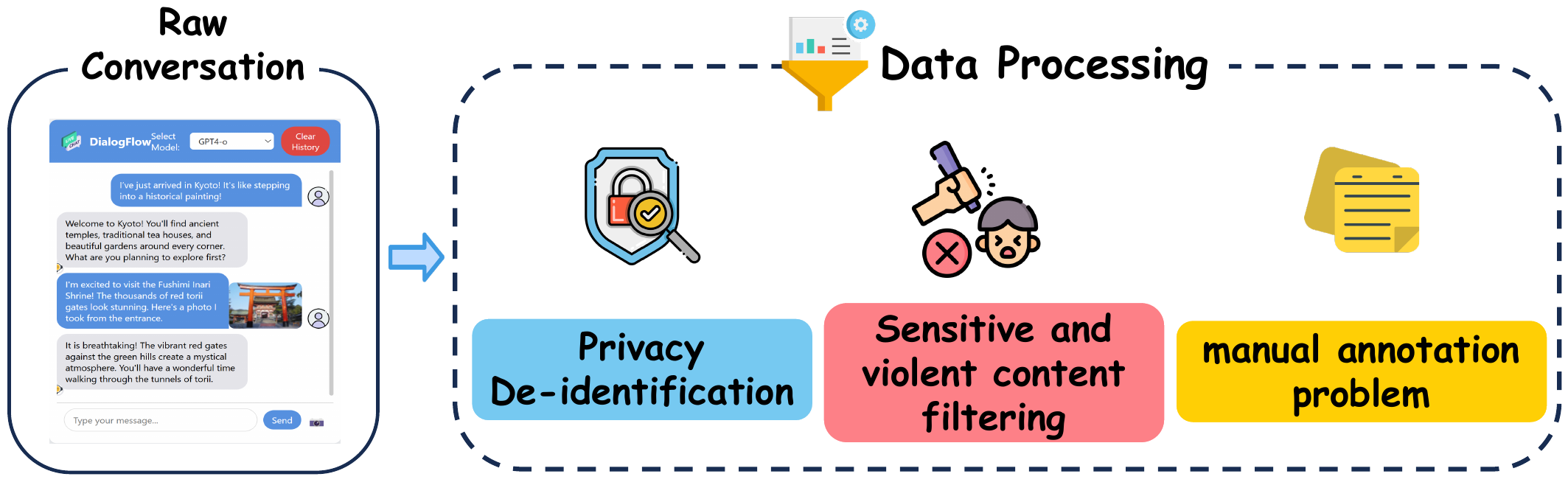} 
  \vspace{-0.2cm}
  \caption{Data construction pipeline of \ours{}.}
  \label{stastic}
  \label{construct}
\end{figure}

 \begin{figure}[!t]  
  \centering
  \includegraphics[width=0.85\linewidth]{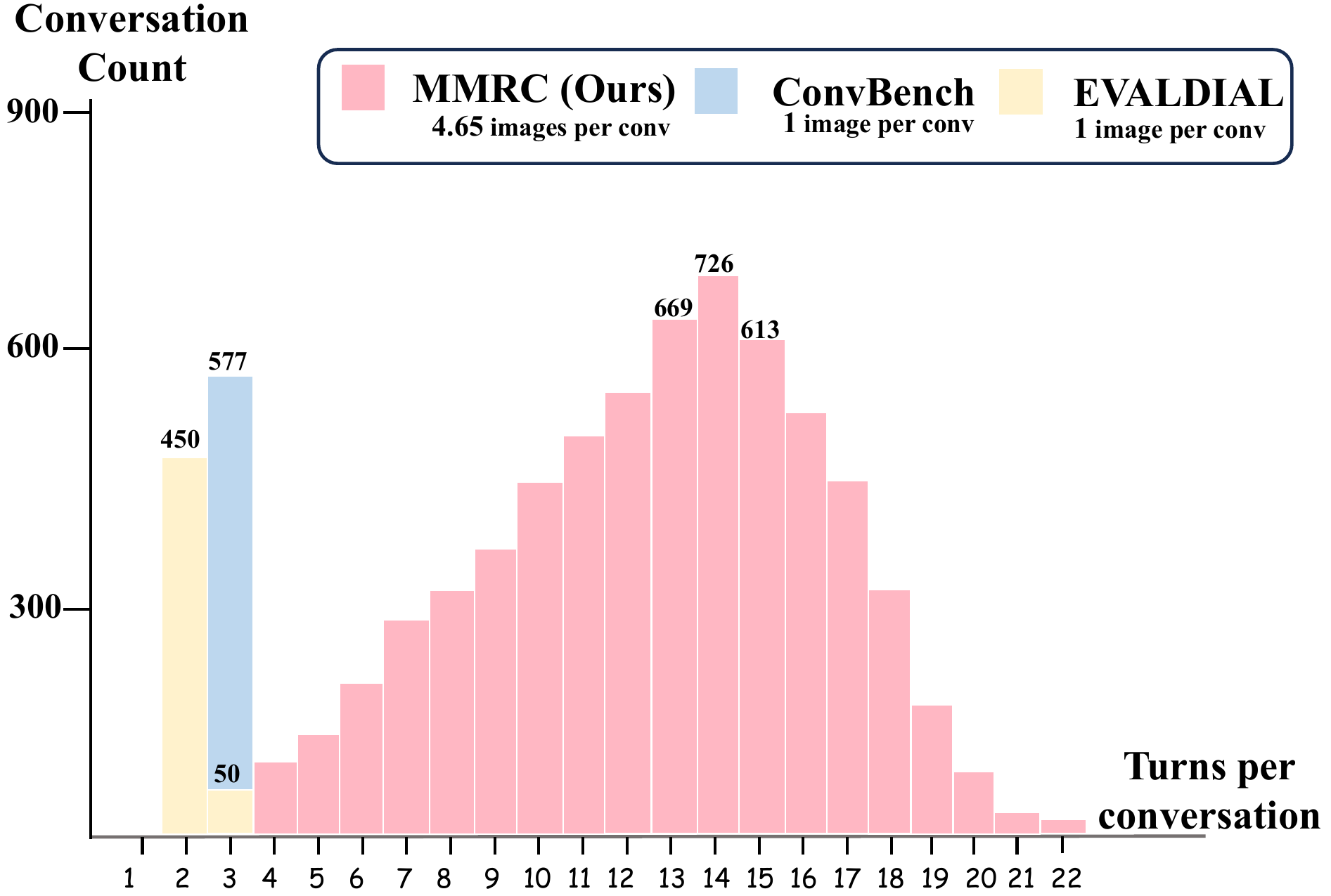} 
  \caption{The distribution of dialogue turns in \ours{}, ConvBench, and EvalDial.}
  \vspace{-0.3cm}
  \label{stastic}
  \label{fig:analyse_turn}
  \vspace{-0.3cm}
\end{figure}

\subsection{Data Curation Process}
We develop {\platform}, a large-scale evaluation platform designed to benchmark 20 cutting-edge MLLMs, with specific models detailed in Appendix~\ref{Dialogflow}. In particular, open-source models are deployed on A100 GPUs, while closed-source models are accessed via APIs. Over an 8-month period, we collected 87,912 raw dialogues from 354 users, leveraging thousands of A100 GPU hours and incurring significant API costs. However, these raw conversations may contain sensitive information, including personal details, violent content, offensive language, biased statements, misinformation, and culturally inappropriate expressions, posing fairness and ethical concerns.

To address this, we design a pipeline to clean the data, as illustrated in Fig.~\ref{construct}. The three stages are as follows: \textit{(i)}: We manually review data for personal information or privacy violations. If any are detected, the relevant segments are deleted, or the entire dialogue is removed if its coherence is compromised. \textit{(ii)}: We also screen for violent, offensive, and sensitive content. If detected, the entire dialogue is removed to prevent the dissemination of harmful material. \textit{(iii)}: We manually annotate clean dialogue data with QA pairs for MLLM evaluation in OEC. These pairs undergo multiple reviews by different annotators to ensure accuracy.

\begin{table}[!t]
\centering
\footnotesize

\vspace{-1em}
\begin{tabular}{@{} l r r @{}}
\toprule
Statistic & Number & Percentage \\
\midrule
Total questions & 28720 & 100\% \\
- Information Extraction  (IE) & 5087 &  17.71\% \\
- Cross-turn Reasoning (CR)& 4789 &  16.67 \% \\
- Information Update (IU)& 4561 &  15.88 \% \\
- Image Management (IM)& 4721 &  16.43\% \\
- Memory Recal (MR)& 4962 &  17.28\% \\
- Answer Refusal (AR) & 4600 &  16.02 \% \\
\midrule
Formats: & & \\
- Open Questions & 24716 & 86.05\% \\
- Multiple-Choice Questions & 2703 & 9.41\% \\
- True/False Questions & 1301 & 4.53\% \\
\midrule
Average image per conversation & 4.65 &  \\
Average conversation length  & 15.2 &  \\
\bottomrule
\end{tabular}
\caption{Problem statistics of \ours.}
\label{tab:data_stastic}
\end{table} 

\begin{figure}
\centering
\vspace{-0.3cm}
  \includegraphics[width=0.85\linewidth]{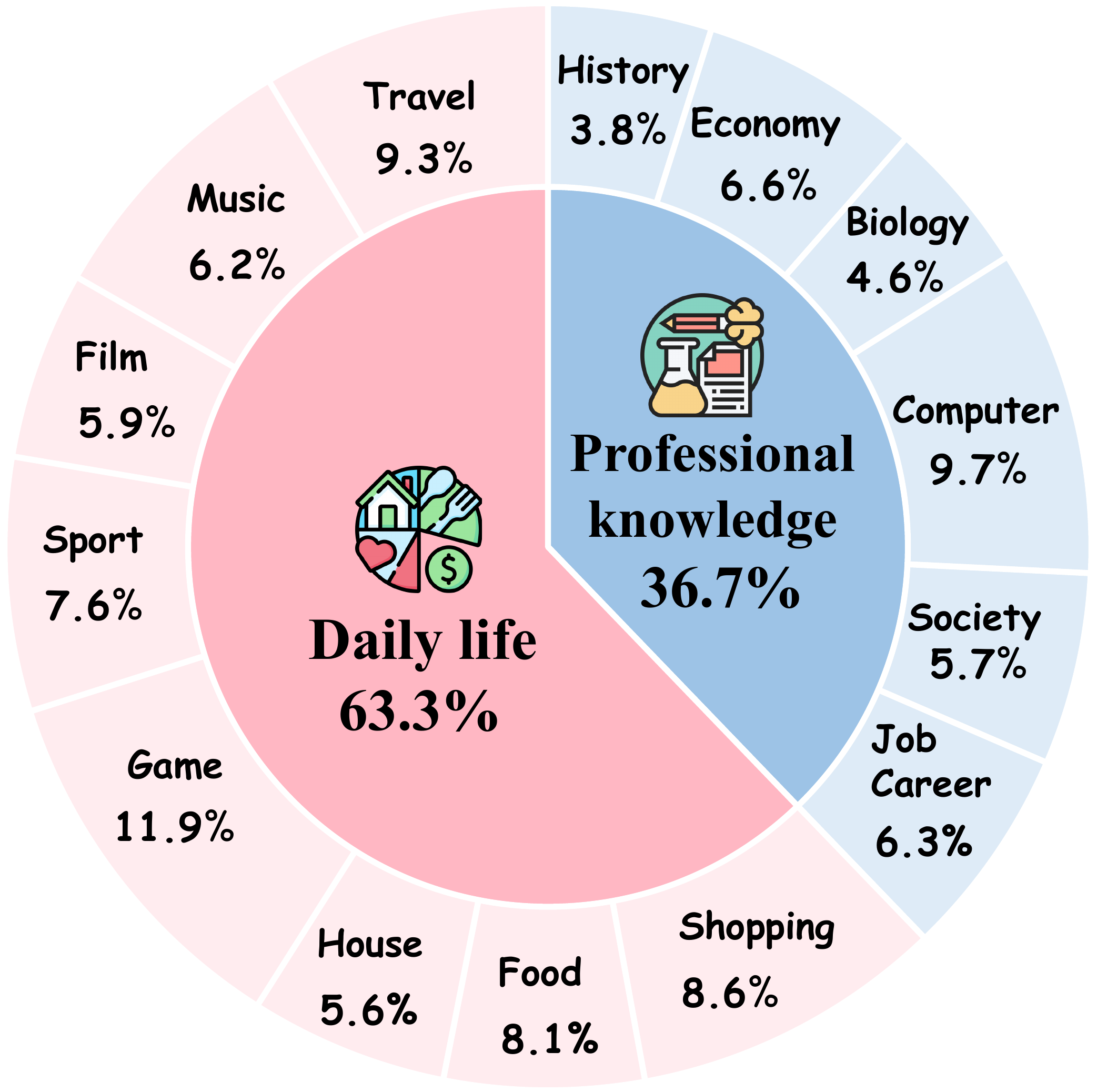}
  \caption{The distribution of conversation categories in our \ours{} dataset.}
  \label{fig:pie}
  \vspace{-0.5cm}
\end{figure}

\subsection{Data Statistics}
We perform a statistical analysis on the distribution of conversation turns, categorization, and questions. The detailed distribution of conversation turns is demonstrated in Fig.~\ref{fig:analyse_turn}. The conversation turns in \ours{} are not fixed, ranging from 4 to 22, making it more natural and realistic compared to the fixed-turn structures in ConvBench~\citep{liu2024convbench} and EVALDIAL~\citep{park2024mitigating}.

To classify diverse real-world conversations, we design a classification network that maps conversation data into 14 predefined categories (details in the Appendix~\ref{Topic net}). The classification results, shown in Fig.~\ref{fig:pie}, indicate that \ours{} exhibit a well-balanced distribution. Moreover, these categories cover a wide range of topics, ensuring the diversity and representativeness of the conversations.

The statistics on manually annotated questions in \ours{} are shown in Table~\ref{tab:data_stastic}. Notably, the number of questions for the six core abilities is well-balanced, with the majority being open-ended questions. Although open-ended questions complicate model evaluation, they provide a finer-grained view of the differences in model responses, enabling a deeper understanding of model performance.

\begin{table*}[t]
\centering
\resizebox{\textwidth}{!}{
\begin{tabular}{llccccccccccccccccc}
    \toprule
    \multirow{2}{*}{\textbf{Type}} & \multirow{2}{*}{\textbf{Model}} & \multicolumn{6}{c}{\textbf{GPT-based Evaluation Metrics}} & \multicolumn{3}{c}{\textbf{Human Evaluation}} & \multicolumn{3}{c}{\textbf{Objective Precision Metrics}} \\ \cmidrule(lr){3-8} \cmidrule(lr){9-11} \cmidrule(lr){12-14}
    & & \textbf{IE} & \textbf{CR} & \textbf{IU} & \textbf{IM} & \textbf{MR} & \textbf{AR}  & \textbf{CR\textsuperscript{*}} & \textbf{IU\textsuperscript{*}} & \textbf{MR\textsuperscript{*}} & \textbf{EP} & \textbf{IMP} & \textbf{RP} & \textbf{Overall} \\
    \midrule
    Human & & 4.79 & 4.78 & 4.82 & 4.87 & 4.77 & 4.80  & 4.81 & 4.79 & 4.66 & 0.974 & 0.981 & 93.26\% & 4.81 \\ \midrule \midrule
    \multirow{16}{*}{Open source} & LLaVA-V1.5-7B & 0.91 & 1.08 & 0.31 & 0.52 & 0.22 & 0.28 & 1.14 & 0.47 & 0.29 & 0.167 & 0.092 & 8.26\%  & 0.55 \\
    & LLaVA-V1.5-13B & 1.04 & 1.21 & 0.25 & 0.92 & 0.26 & 0.15  & 1.32 & 0.42 & 0.33 & 0.206 & 0.193 & 6.14\%  & 0.64\\
    & MiniCPM-8B & \textbf{4.08} & 3.94 & 2.98 & \textbf{3.47} & 3.65 & \textbf{3.78}  & 4.08 & 2.91 & 3.79 & \textbf{0.748} & 0.654 & \textbf{79.23\%} & \underline{3.65} \\
    & LLaVA-Next-0.5B & 2.32 & 2.89 & 1.99 & 1.87 & 2.67 &1.12 & 2.63 & 2.11 & 2.73 & 0.446 & 0.358 & 20.58\% & 2.14 \\
    & LLaVA-Next-7B & 3.23 & 3.85 & 2.77 & 2.18 & \underline{4.01} &  2.08 & 2.92 & 2.65 & \textbf{4.02} & 0.611 & 0.355 & 36.21\% & 3.02  \\
    & Qwen2VL-2B & 2.16 & 2.85 & 1.41 & 1.27 & 1.93 & 2.33 & 2.71 & 1.62 & 2.23 & 0.336 & 0.269 & 40.29\%  & 1.99  \\
    & Qwen2VL-7B & 2.82 & 3.68 & 2.62 & 2.16 & 2.11 & 1.57  & 3.72 & 2.77 & 2.19 & 0.531 & 0.408 & 30.62\% & 2.49 \\
    & Qwen2VL-72B & 3.17 & \textbf{4.17} & 2.78 & 2.73 & 2.81 & 1.32 & \textbf{4.15} & 2.81 & 2.89 & 0.603 & 0.524 & 25.41\% & 2.83 \\
    & LLaVA-OneVision-0.5B & 2.03 & 3.01 & 2.44 & 1.79 & 2.55 & 1.92  & 3.11 & 2.40 & 2.57 & 0.305 & 0.436 & 36.85\% & 2.29 \\
    & LLaVA-OneVision-7B & 3.52 & 3.71 & \underline{3.21} & 2.29 & 3.35 & \underline{2.93} & 3.82 & 3.29 & 3.33 & 0.653 & 0.439 & \underline{66.23\%} & 3.16  \\
    & LLaVA-OneVision-72B & \underline{4.06} & \underline{4.08} & \textbf{4.01} & 3.24 &
    \textbf{4.17} & 2.52 & \underline{4.11} & \textbf{3.89} & \underline{3.99} & \underline{0.723} & \textbf{0.677} & 54.28\% & \textbf{3.68}  \\
    & VILA1.5-3B & 3.08 & 2.79 & 2.87 & 2.91 & 3.42 & 2.23  & 2.83 & 3.08 & 3.30 & 0.566 & 0.502 & 47.97\% & 2.88 \\
    & VILA1.5-8B & 3.46 & 3.45 & 3.12 & 2.84 & 3.66 & 2.74  & 3.61 & \underline{3.34} & 3.67 & 0.659 & 0.542 & 59.21\% & 3.22 \\
    & mplug-Ow3-1B & 2.48 & 2.76 & 1.92 & 2.53 & 2.78 & 1.51  & 2.61 & 2.08 & 2.77 & 0.503 & 0.547 & 32.27\% & 2.33 \\
    & mplug-Ow3-2B & 3.45 & 2.91 & 2.36 & 2.71 & 2.42 & 2.09  & 2.99 & 2.41 & 2.40 & 0.522 & 0.610 & 42.14\% & 2.66 \\
    & mplug-Ow3-7B & 3.92 & 3.89 & 2.59 & \underline{3.34} & 2.83 & 2.91  & 3.95 & 2.78 & 2.81 & 0.702 & \underline{0.672} & 62.49\% & 3.25 \\ \midrule
    & Avg. & 2.86 & 3.14 &2.35 &2.30 &2.68& 1.97 & 3.11& 2.44 &2.71& 0.518& 0.455 & 40.89\% & 2.55\\
    \midrule
    \midrule
    \multirow{4}{*}{Closed source} & GPT-4o & \textbf{4.35} & \textbf{4.38} & \textbf{4.28} & \textbf{4.12} & \textbf{4.31} & 3.06 & \textbf{4.26} & \textbf{4.16} & \textbf{4.18} & \textbf{0.905} & \textbf{0.826} & 68.28\%  &\textbf{4.08}  \\
    & Claude-3.5-sonnet & \underline{4.12} & \underline{4.04} & \underline{3.98} & 3.89 & 3.88 & \underline{3.21}  & \underline{4.09} & \underline{4.07} & 3.92 & \underline{0.823} & 0.786 & \underline{74.28\%} & \underline{3.86} \\
    & Gemini-1.5 Pro & 3.96 & 3.90 & 3.75 & 3.94 &  \underline{4.16} & 3.11  & 3.92 & 3.92 & \underline{4.10} & 0.716 & 0.794 & 73.03\%  & 3.80\\
    & DeepSeek-V3 & 3.92 & 3.96 & 3.92 & \underline{3.98} & 4.03 & \textbf{3.28} & 4.01 & 3.91 & 3.99 & 0.702 & \underline{0.798} & \textbf{74.93\%}  & 3.84\\
    \midrule
    & Avg. & 4.09 & 4.07 & 3.98 & 3.99& 4.10&3.17 & 4.07&4.01 &4.04 & 0.787& 0.801 & 72.63\% & 3.90\\
    \bottomrule
\end{tabular}}
\caption{Comparison of Performance for 20 MLLMs on \ours{}. \textasteriskcentered\ indicates that the same task has been re-evaluated manually. \textbf{Bold} and \underline{underline} denote the best and second-best results, respectively.}
\label{tab:result}
\vspace{-0.5cm}
\end{table*}

\vspace{-0.2cm}
\section{Experiment and Analysis}
\vspace{-0.1cm}
\subsection{Evaluation Matrix}

\begin{figure}[!t]
\centering
\includegraphics[width=\linewidth]{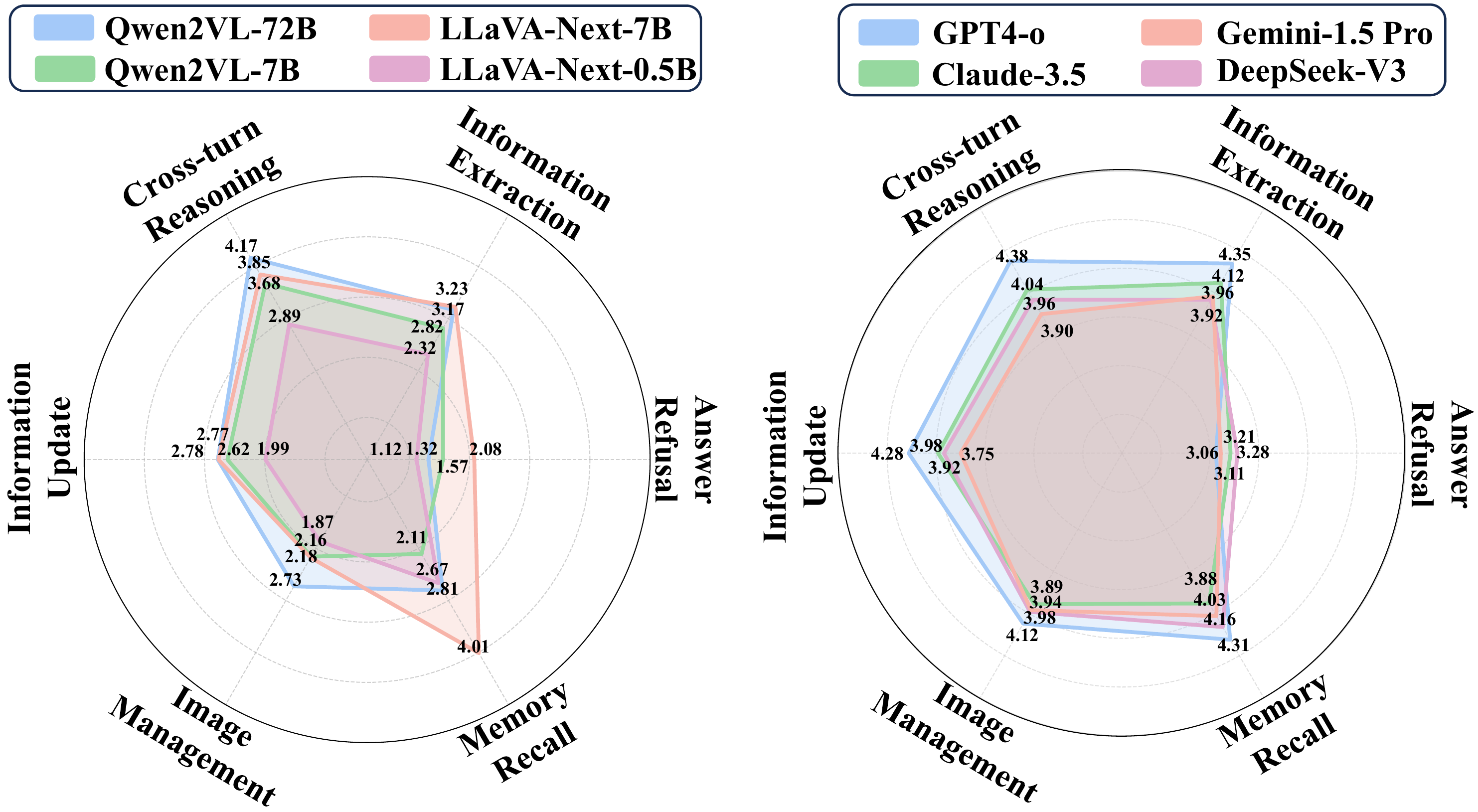}
\vspace{-0.4cm}
\caption{Radar chart of capabilities for models with noticeable task-specific imbalances.}
\label{fig:radar2}
\vspace{-0.5cm}
\end{figure}

Since questions in \ours{} are open-ended, directly evaluating accuracy is infeasible. To address this, we develop a comprehensive evaluation framework that integrates GPT-based scoring, human assessment, and objective precision metrics. Specifically, GPT evaluates all six abilities (Section~\ref{3.1}), while human evaluators conduct a second-round review for CR, IU, and MR, which require more in-depth judgment. Moreover, both GPT and human evaluators employ a scoring scale ranging from 0 to 5, with prompts and evaluation criteria detailed in the Appendix~\ref{Prompt templete}. In contrast, for IE, IM, and AR, we employ objective precision metrics, including Extraction Precision (EP), Image Management Precision (IMP), and Refuse Precision (RP), to provide an intuitive assessment of model performance by measuring the proportion of correct responses.

\begin{figure*}[h]
  \centering
  \vspace{0.3cm}
  \begin{minipage}{0.33\textwidth}
    \centering
    \includegraphics[width=0.92\linewidth]{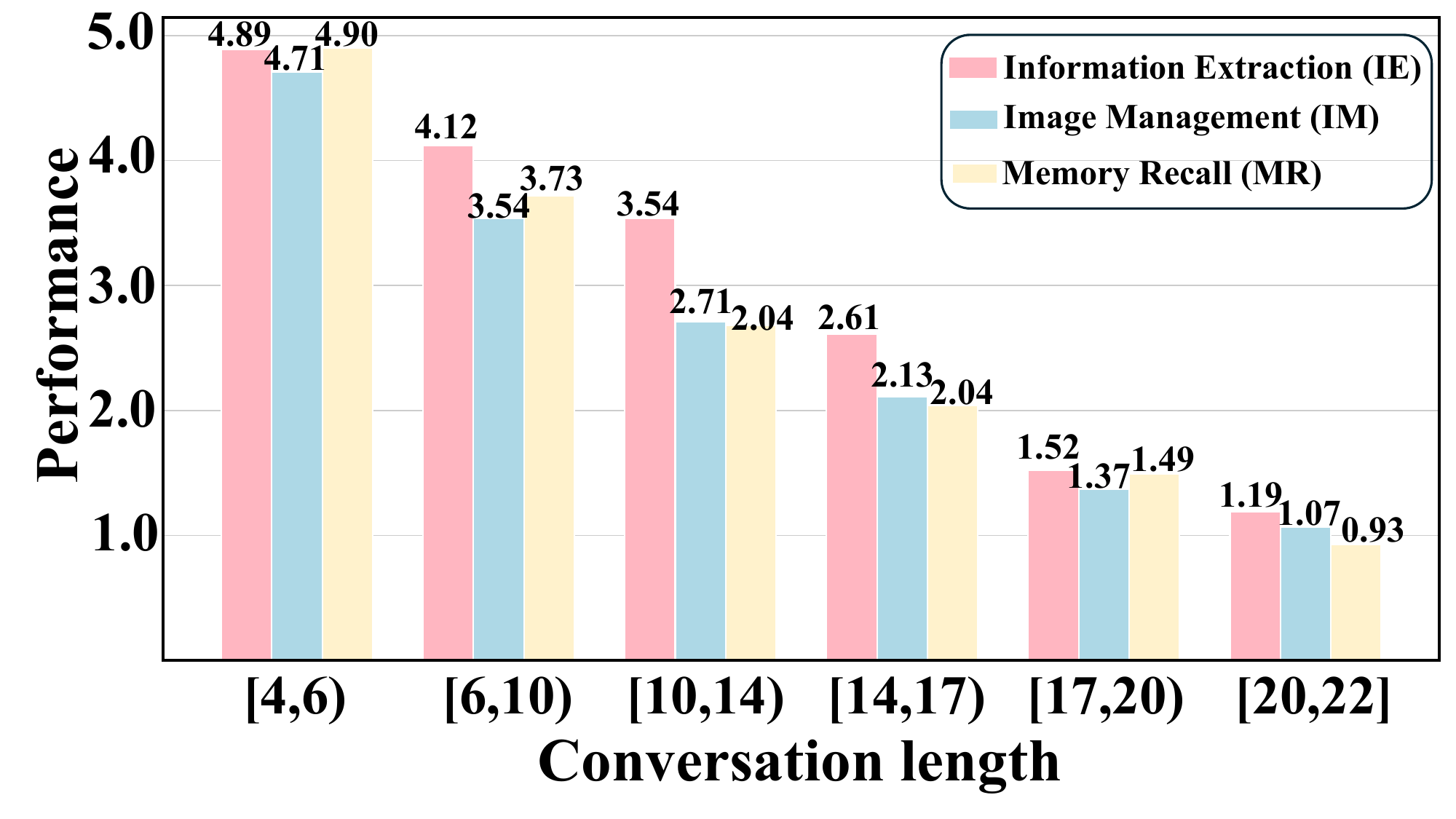}
    \caption{The impact of conversation length on memory performance.}
    \label{fig:turn_comperation}
  \end{minipage}
  \hfill
  \begin{minipage}{0.32\textwidth}
    \centering
        \vspace{0.2em}
    \includegraphics[width=\linewidth]{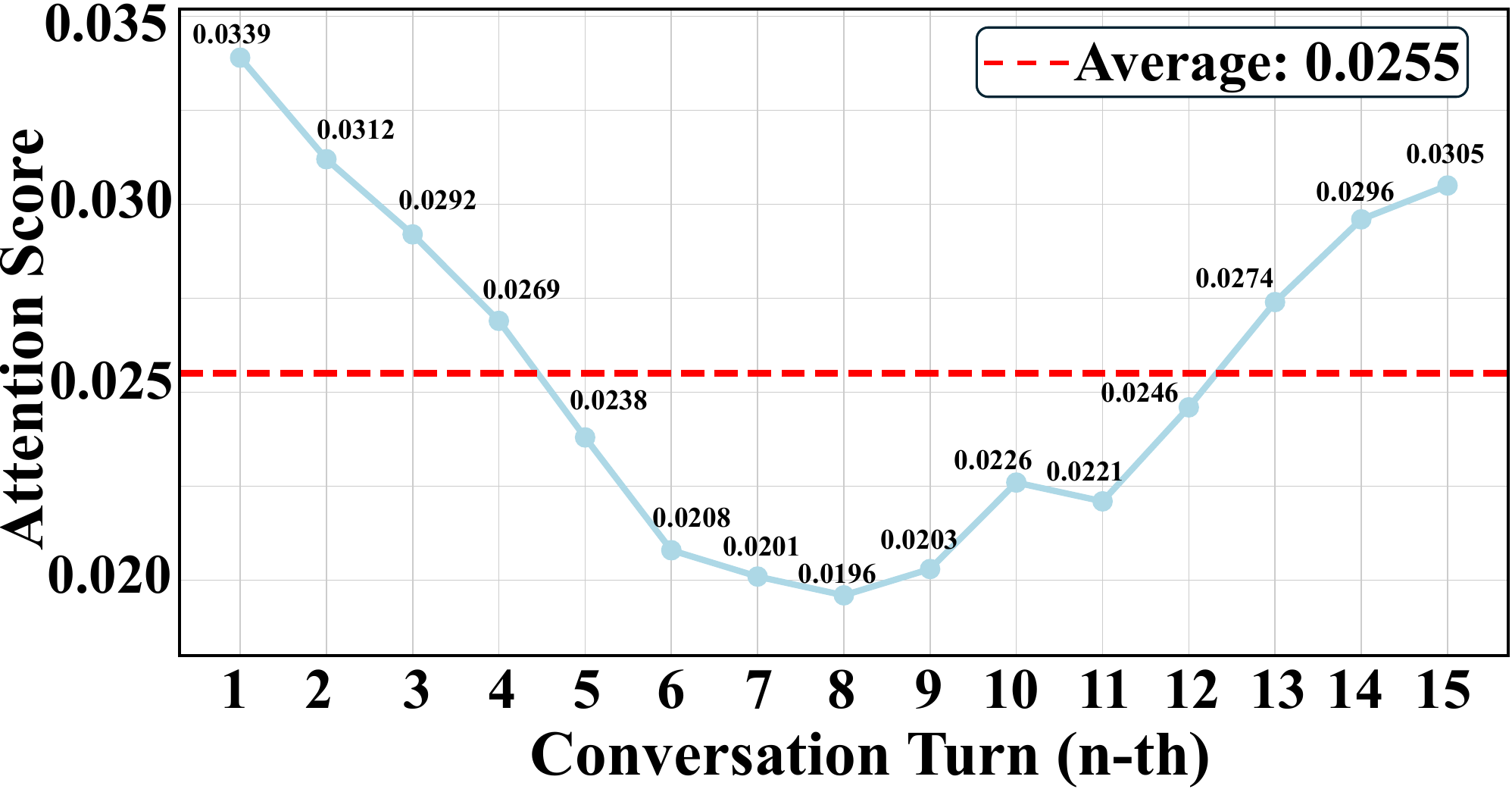}
    \caption{Different turns' attention score in the conversation.}
    \label{fig:attention}
  \end{minipage}
  \hfill
  \begin{minipage}{0.32\textwidth}
    \centering
    \includegraphics[width=\linewidth]{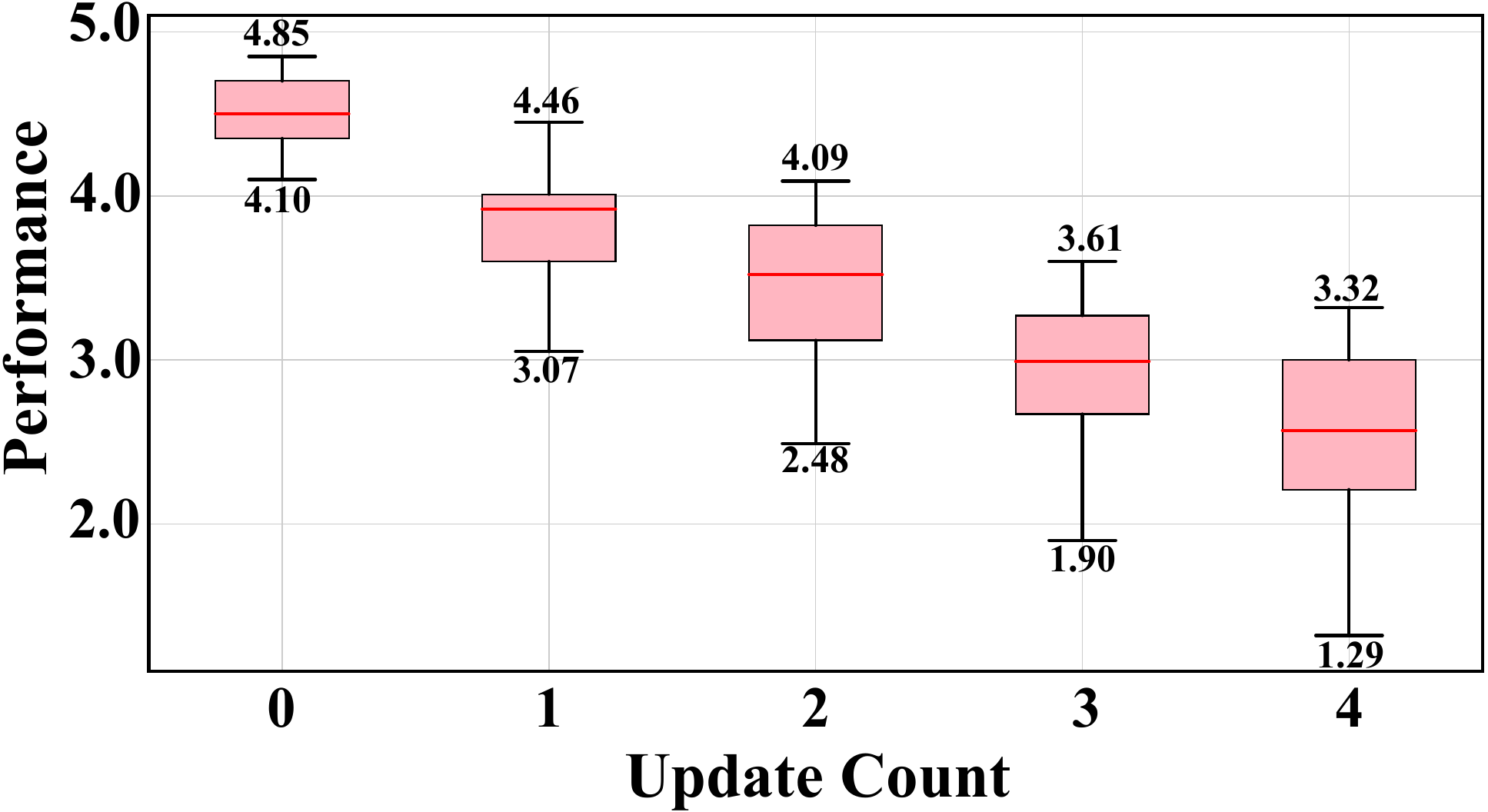}
    \caption{Impact of update frequency on model performance.}
    \label{fig:update}
  \end{minipage}
  \vspace{-12pt}
\end{figure*}

\noindent\textbf{EP as an extension of IE, measures the precision in extracting items:} 

\[
\text{EP} = \frac{|\text{Model}_{\text{items}} \cap \text{Label}_{\text{items}}|}{|\text{Model}_{\text{items}}|},
\]
where $\text{Model}_{\text{items}}$ denotes the set of items generated by the model, and $\text{Label}_{\text{items}}$ denotes the set of ground-truth items.

\noindent\textbf{IMP as an extension of IM, measures the precision in managing and retrieving images:} 

\[
\text{IMP} = \frac{|\text{Image}_{\text{hit}}|}{|\text{Image}_{\text{hit}} \cup \text{Image}_{\text{miss}}|},
\]
where \(\text{Image}_{\text{hit}}\) denotes the images correctly retrieved by the model, and \(\text{Image}_{\text{miss}}\) denotes the images that are part of the correct answer but were not retrieved by the model.

\noindent\textbf{RP as an extension of AR, measures the precision in refusing to answer unknown questions:} 
\[
\text{RP} = \frac{\sum_{i=1}^{N} \mathbbm{1}(D_i)}{\sum_{i=1}^{N} \mathbbm{1}(E_i)},
\]
where \( D_i \) and \( E_i \) denote the model's refusal to answer and the ground-truth refusal for the \( i \)-th question, respectively. The indicator function \( \mathbbm{1}(\cdot) \) returns 1 if the response aligns with the expected refusal behavior and 0 otherwise. \( N \) denotes the total number of questions.

\subsection{Main Results}
Table~\ref{tab:result} presents the performance of 20 open-source and closed-source MLLMs in real-world dialogue scenarios. Based on the evaluation results, we identify three key findings:

\noindent \textbf{(1) Challenges of Reality:} LLaVA-1.5 performs poorly in OEC, while powerful GPT-4o still falls short of human-level performance in real-world scenarios, highlighting the complexity and difficulty of practical, user-driven conversations in \ours{}.

\noindent \textbf{(2) Reliability of Evaluation:} The similarity between GPT’s scores and those of human annotators reaches 93\%. Furthermore, GPT's scores closely correspond with the Objective Precision Metrics, where higher GPT scores consistently reflect stronger performance. This consistency indicates the effectiveness of GPT guided by our evaluation prompts, providing a foundation for using GPT scores in subsequent experimental analysis.

\noindent \textbf{(3) Task-Specific Imbalance:} 
The performance of most models is imbalanced, exhibiting distinct strengths and weaknesses. As illustrated in Fig.~\ref{fig:radar2}, the LLaVA-Next family demonstrates strong memory recall ability but weaker image management ability. Similarly, the Qwen2VL family excels in cross-turn reasoning yet exhibits a relatively weak answer refusal ability. Notably, closed-source models exhibit a more pronounced performance imbalance, consistently struggling with answer refusal. This disparity in MLLMs stems from variations in training datasets and strategies. Specifically, different organizations prioritize fine-tuning for certain tasks, leading to enhanced performance in those areas while resulting in weaker performance on tasks with less targeted training.

\subsection{Error Analysis}

We conduct an in-depth analysis of failures of the model and identify four common error patterns:

\noindent\uline{(1) Long-term memory degradation:} As the conversation progresses, the model's memory of previous dialogue content becomes increasingly vague, a phenomenon known as memory degradation~\cite{zhong2024memorybank}. Moreover, the severity of memory degradation increases as the conversation lengthens. As illustrated in Fig.~\ref{fig:turn_comperation}, memory-related abilities (\textit{i.e.,} IE, IM, MR) decline significantly in extended conversations. Observations reveal that memory degradation is more severe in the middle of a conversation than at the beginning or end, challenging the assumption that earlier memories degrade more rapidly. To further investigate, we visualize the model's attention patterns when addressing memory-related questions (Appendix~\ref{Attention Calculation}). As shown in Fig.~\ref{fig:attention}, attention to the middle part of the conversation is markedly lower than to the beginning and end, mirroring the observed degradation pattern. We hypothesize that this unbalanced attention distribution contributes to the heightened memory degradation in mid-conversation.

\begin{figure*}[!t]
  \centering
  \vspace{-0.3cm}
  \begin{minipage}{0.25\textwidth}
    \centering
    \vspace{-0.5cm}
    \includegraphics[width=0.90\linewidth]{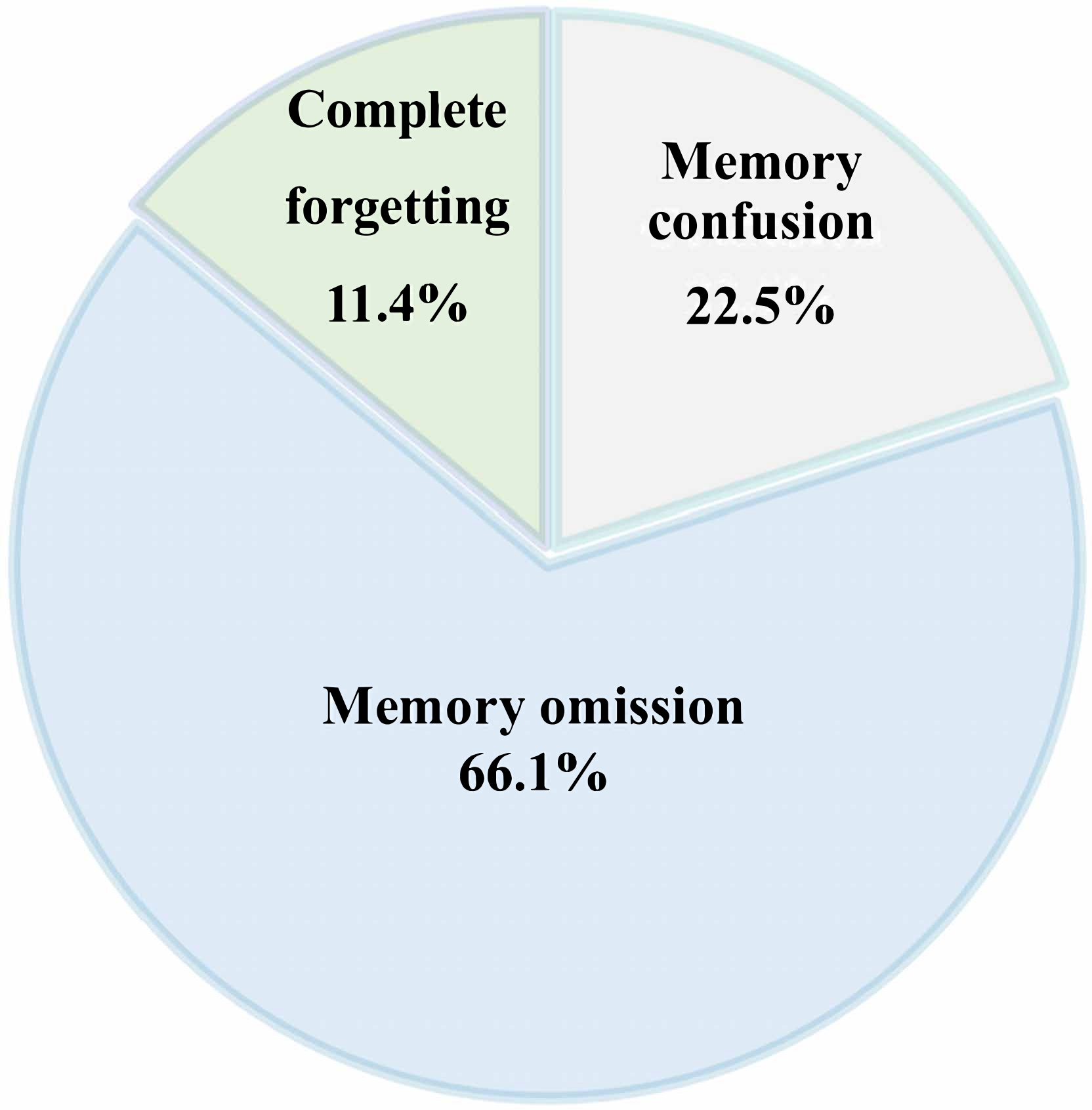}
    \vspace{0.6em}
    \vspace{-0.5cm}
    \caption{Information extraction error statistics.}
    \label{fig:IE_pie}
  \end{minipage}
  \hfill
  \begin{minipage}{0.25\textwidth}
    \centering
     \vspace{-0.5cm}
    \includegraphics[width=0.90\linewidth]{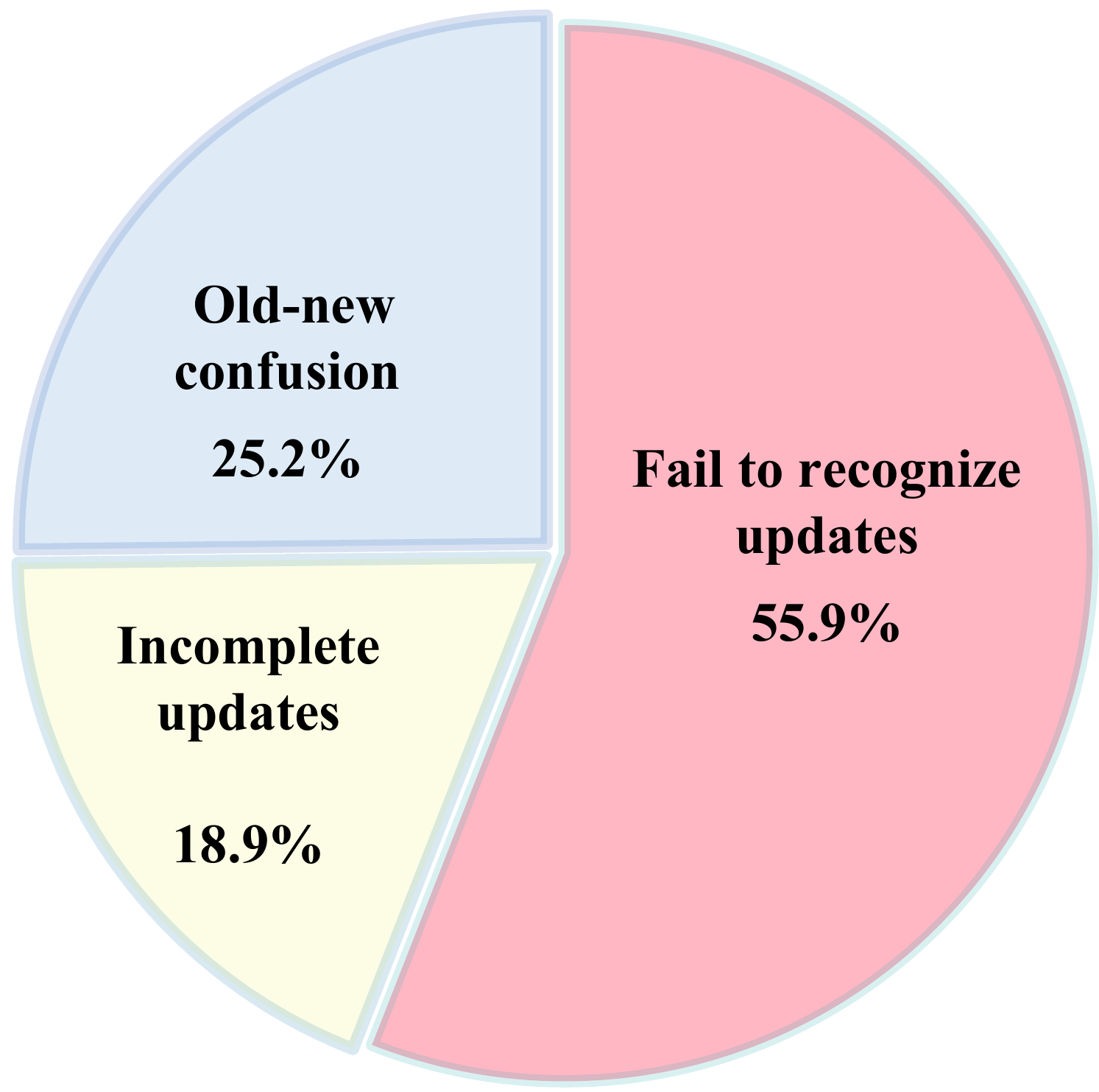}
    \vspace{0.6em}
    \vspace{-0.4cm}
    \caption{Information update error statistics.}
    \label{fig:update_pie}
  \end{minipage}
  \hfill
  \begin{minipage}{0.23\textwidth}
    \centering
        \vspace{-1.0em}
    \includegraphics[width=1.05\linewidth]{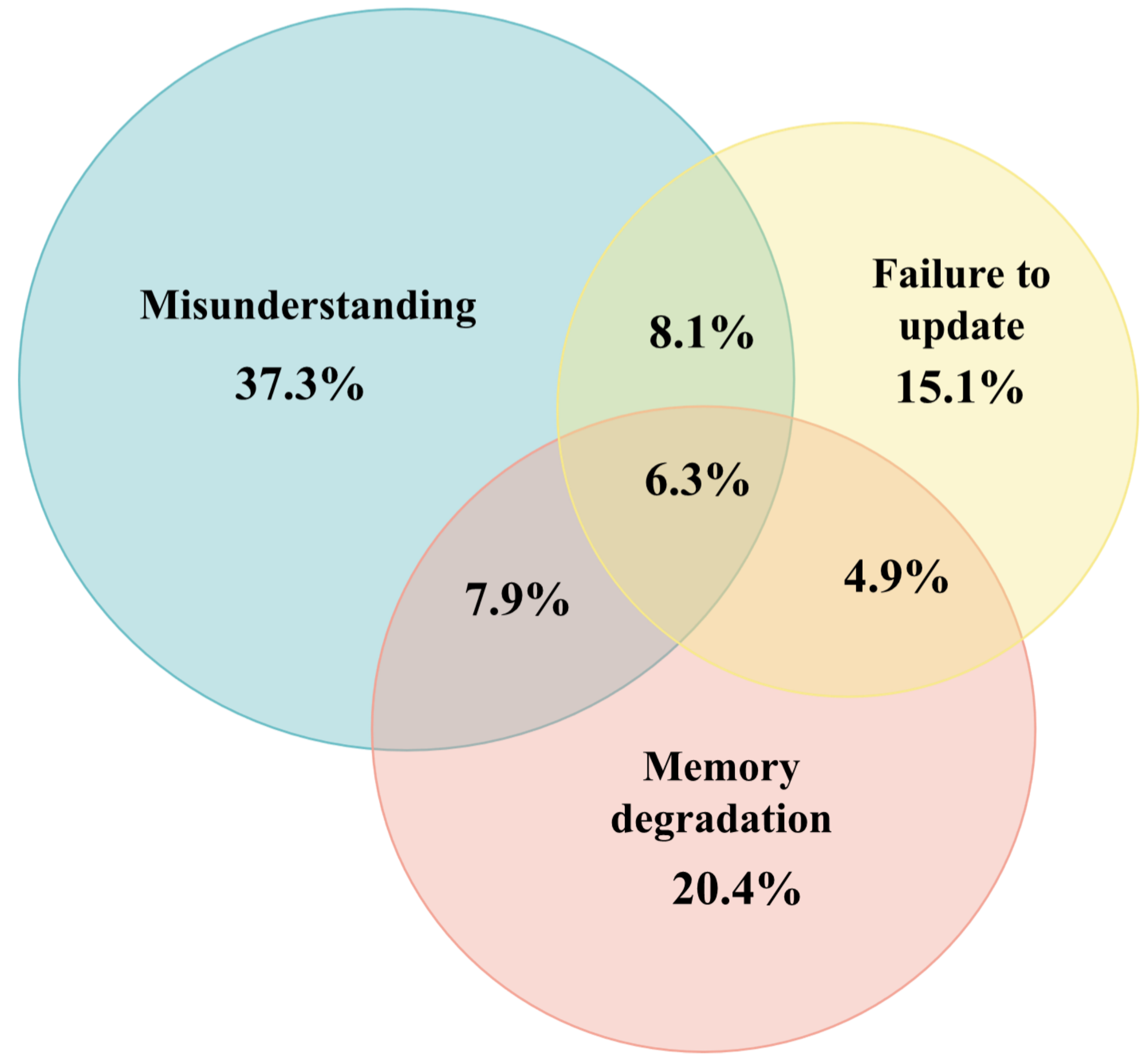}
    \vspace{-0.7em}
    \vspace{-0.5cm}
    \caption{Statistics of error types in reasoning.}
    \label{fig:cr_pie}
  \end{minipage}
  \hfill
    \begin{minipage}{0.23\textwidth}
    \centering
    \vspace{-1.05em}
    \includegraphics[width=0.90\linewidth]{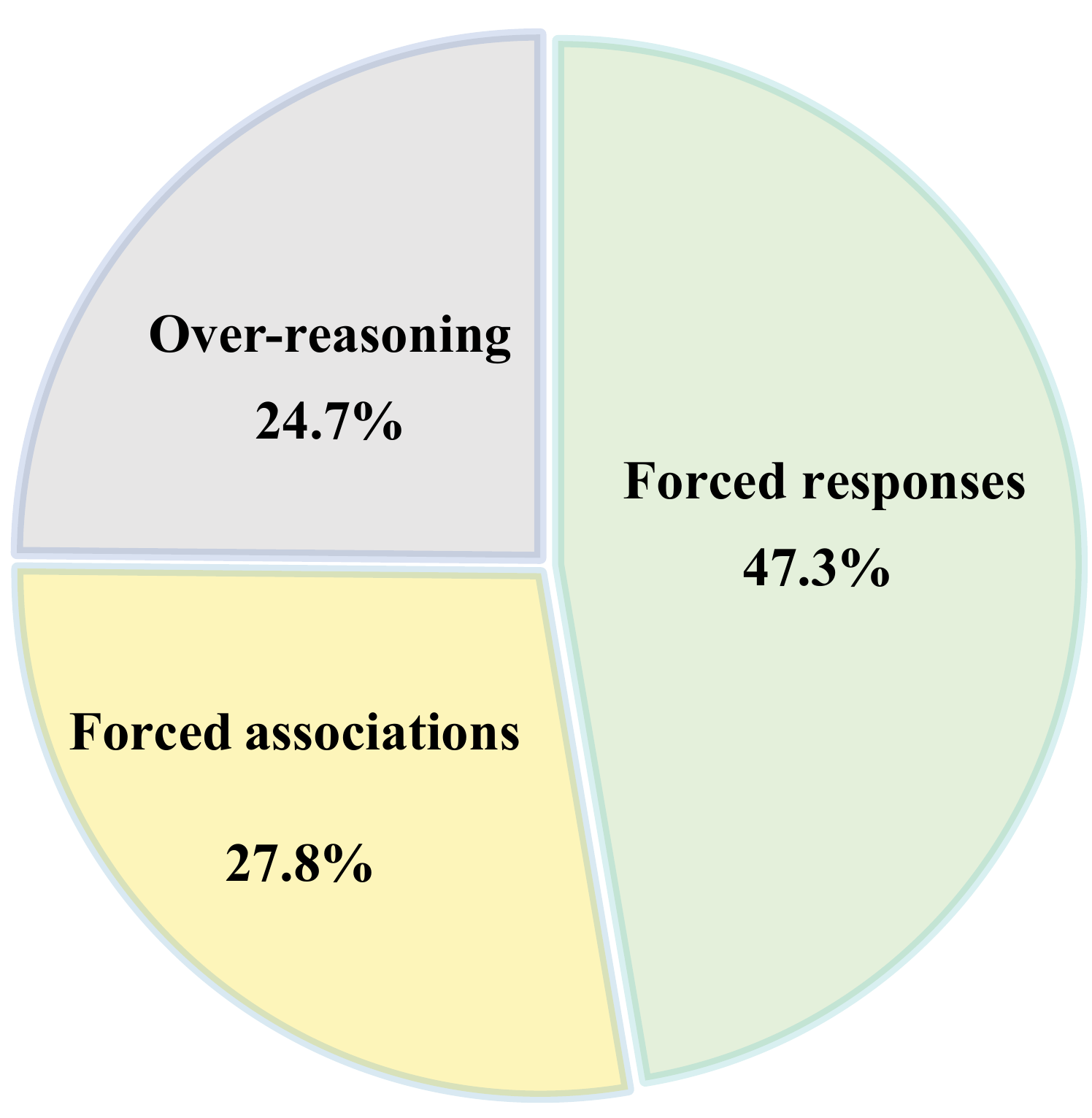}
    \vspace{1.0em}
    \vspace{-0.5cm}
    \caption{Statistics of error types in answer refusal.}
    \label{fig:ar_pie}
  \end{minipage}
\vspace{-0.5cm}
\end{figure*}

Furthermore, we analyze the error types associated with long-term memory degradation, with results shown in Fig.~\ref{fig:IE_pie}: \textit{(i) Memory omission:} the model fails to retain certain details from the conversation. \textit{(ii) Complete forgetting:} a more severe form of memory degradation than omission, where the model entirely fails to recall specific events from the conversation. \textit{(iii) Memory confusion:} the model incorrectly merges relevant information with unrelated content, leading to distorted recollection.

\begin{figure}[!t]
\centering
\vspace{0.7em}
\includegraphics[width=0.95\linewidth]{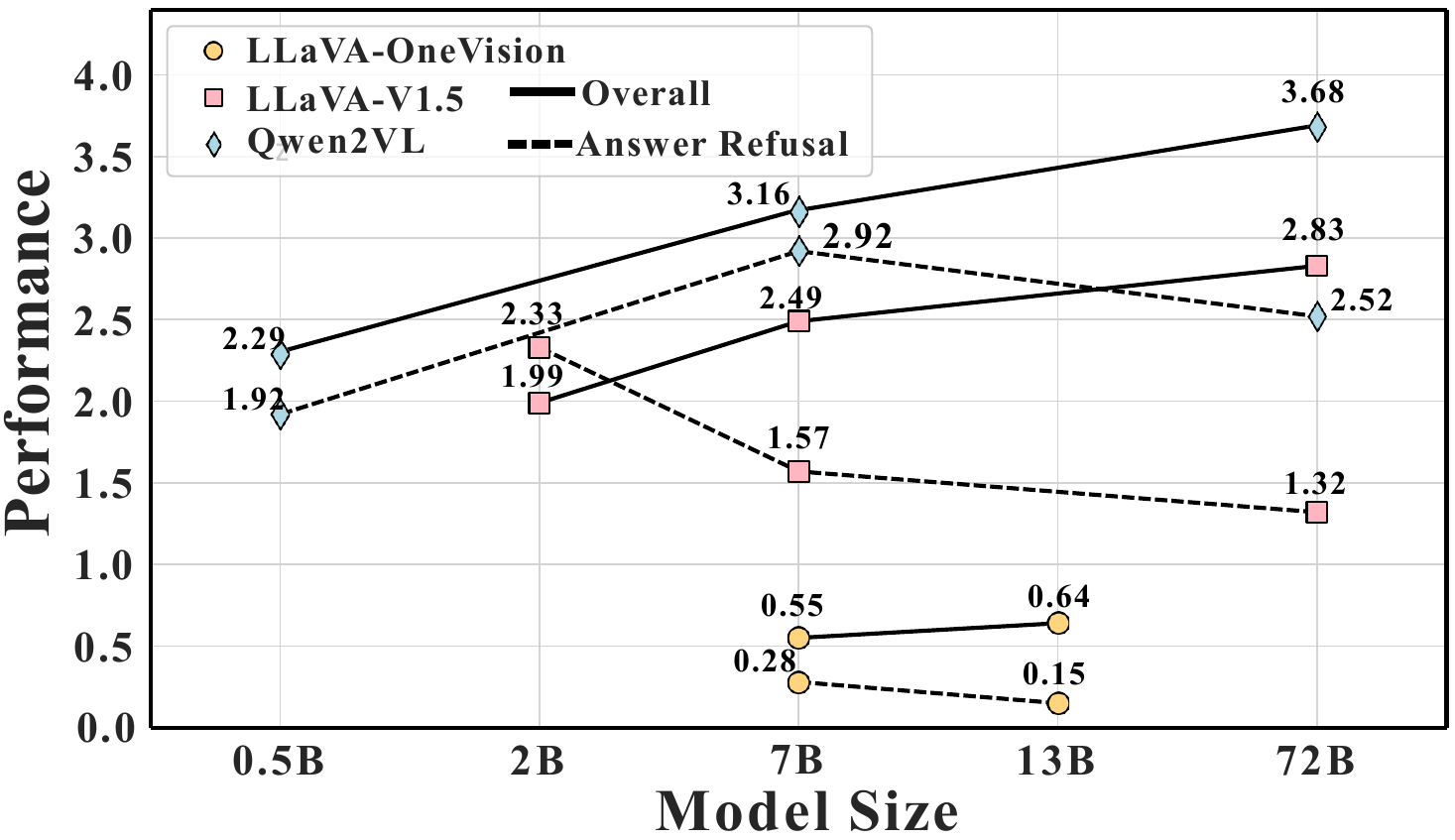}
\vspace{-0.25cm}
\caption{Overall and answer refusal performance across different model sizes within the same family.}
\label{fig:size}
\vspace{-0.519cm}
\end{figure}

\vspace{0.1cm}
\noindent\uline{(2) Inadequacies in updating factual knowledge:} MLLMs often struggle to track changes in user information and factual knowledge during conversations, resulting in failures in updating information. As illustrated in Fig.~\ref{fig:update}, frequent changes in factual knowledge make it difficult to adapt to rapidly evolving information and result in a decline in update performance. To further investigate this issue, we analyze the types of errors associated with updating information, with their distribution shown in Fig.~\ref{fig:update_pie}: \textit{(i) Failure to recognize updates:} occurs when the model fails to detect that certain factual knowledge requires updating, instead treating it as static information. \textit{(ii) Incomplete updates:} arises when the model acknowledges the need for an update but fails to incorporate the most recent information due to frequent changes. \textit{(iii) Old-new contradiction:} happens when the model incorrectly merges outdated facts with new ones, leading to an inaccurate representation of the latest information.

\vspace{0.1cm}
\noindent\uline{(3) Accumulated assumption of error propagation:}
During reasoning, the model processes information sequentially from earlier to later dialogue turns to comprehend the dialogue and integrate relevant details, forming a reasoning chain to answer complex questions. However, as illustrated in Fig.~\ref{fig:cr_pie}, the formation of this reasoning chain is often hindered by three key issues, leading to flawed reasoning and incorrect answers: \textit{(i) Misunderstanding:} the model fails to correctly understand the dialogue content, resulting in distortions within the reasoning chain and incorrect assumptions that ultimately lead to erroneous conclusions. \textit{(ii) Memory degradation:} the model forgets prior conversation information, disrupting the reasoning chain. This lack of essential context weakens the model’s assumptions, preventing accurate reasoning. \textit{(iii) Failure to update:} the model continues reasoning based on outdated knowledge, leading to incorrect answers. Notably, these errors can occur simultaneously. As the conversation progresses, their accumulation exacerbates incorrect assumptions, causing errors to propagate more severely throughout the dialogue.

\vspace{0.15cm}
\noindent\uline{(4) Reluctance to “say no.”:}
The model provides unreliable answers when the context is insufficient, potentially misleading users. To understand the underlying causes of this issue,  we conduct an analysis and categorize them into three key types, as illustrated in Fig.~\ref{fig:ar_pie}. \textit{(i) Forced Responses:} the model recognizes that the question is unrelated to the given context and does not utilize any conversational context for its response, yet it fails to refuse to answer. \textit{(ii) Over-reasoning:} the model excessively analyzes the question, attempting to infer non-existent details from the available context. \textit{(iii) Force associations:} the model artificially links an irrelevant element in the question with existing conversation details, generating an answer based on this false connection. Furthermore, as shown in Fig.~\ref{fig:size}, within the same model family, larger models demonstrate superior overall performance. However, their ability to refuse inappropriate answers declines. 
A comparative analysis with varying sizes reveals that while larger models exhibit stronger logical reasoning capabilities, they are more prone to over-reasoning and forced associations compared to smaller models.

\begin{figure}[!t]
\centering
\includegraphics[width=0.97\linewidth]{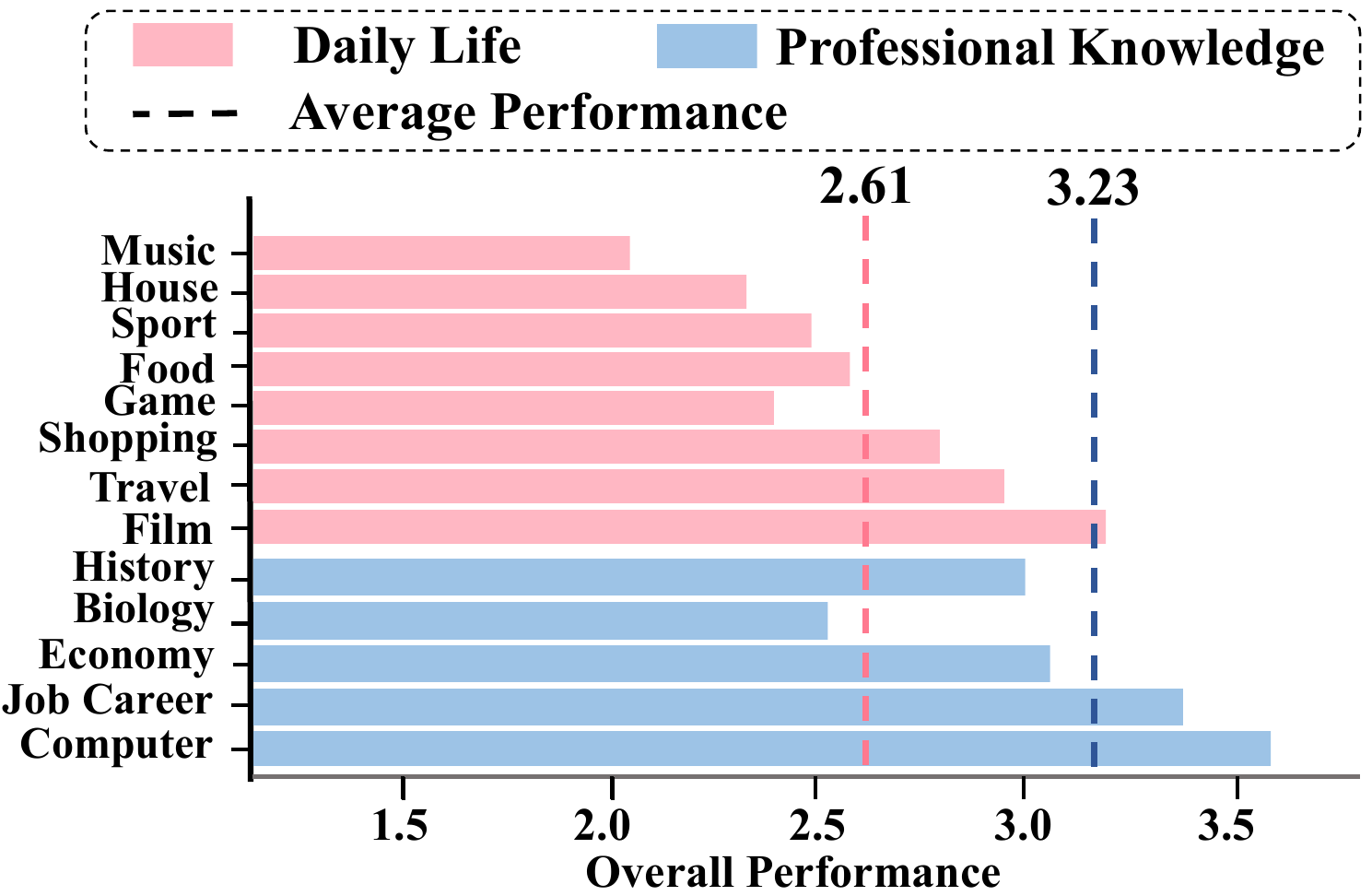}
\vspace{-0.5em}
\caption{MLLMs' performance in different topics.}
\label{fig:domain_bias}
\vspace{-0.5cm}
\end{figure}

\vspace{-0.2cm}
\subsection{Further Discussion}

\noindent\textbf{Domain Bias:} 
To further investigate the model's performance across different domains, we evaluate its overall effectiveness in each domain, as shown in Fig.~\ref{fig:domain_bias}. Our analysis reveals that the model performs significantly better in professional knowledge conversations than in daily conversations (professional: 3.23, daily: 2.61). We hypothesize that this disparity stems from variations in the instruction-based training across models. Specifically, models demonstrating stronger performance in professional knowledge conversations benefit from a larger proportion of instruction-based data tailored for knowledge-based tasks. Thus, to improve MLLMs' conversational abilities in OEC, it is essential to incorporate more daily conversation data during supervised fine-tuning.

\vspace{0.1cm}
\noindent\textbf{Modalities Preference:} To explore the model's preference for different information modalities, we modify 100 conversations by replacing parts of the original text content with equivalent image inputs. For instance, the text-based statement: “I visited the Eiffel Tower.” is converted into “I visited this place.” followed by an image of the Eiffel Tower. The rest of the dialogue remains unchanged for evaluation. Our findings indicate that MLLMs exhibit a strong preference for text-based information, with overall scores for text-based dialogues being 26.3\% higher than their image-based counterparts. Furthermore, models exhibit fewer memory degradation errors in text-based conversations, as memory-related capabilities such as IE, IM, and MR show a 34.6\% improvement. We attribute this to two main factors: \textit{(i):} images often require more tokens to convey the same meaning as text, significantly increasing context length. \textit{(ii):} The model's training data is imbalanced, with text data vastly exceeding image data, leading to stronger proficiency in processing textual information.

\vspace{-0.2cm}
\section{\strategy{} as Improved Baseline}
\vspace{-0.1cm}

\label{sec:ours_method}

\begin{figure}
    \centering
    \includegraphics[width=0.95\linewidth]{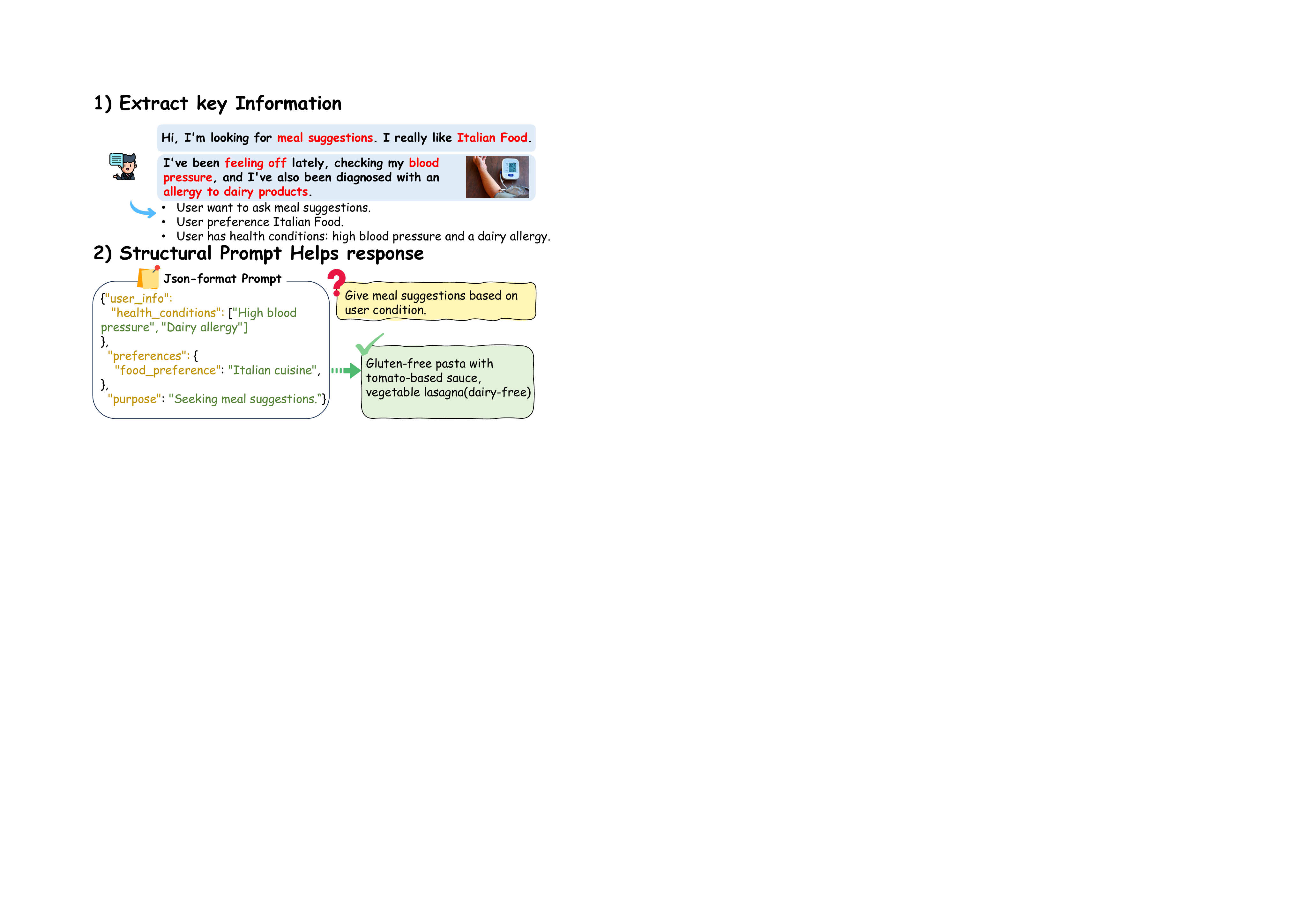}
    \vspace{-0.2cm}
    \caption{\textbf{Illustration of \strategy{} method.}}
    \label{fig:notetaking}
    \vspace{-0.4cm}
\end{figure}

In this section, we introduce an initial step toward enhancing the core capabilities of MLLMs in OEC. The primary failure of MLLMs is their inability to accurately retrieve detailed information, update knowledge, and recognize missing information. To mitigate this, we propose a \strategy{} framework, which guides MLLMs in extracting key dialogue information and recording it in accessible json-format notes. These structured notes serve as external memory, improving response accuracy and context understanding.

As illustrated in Fig.~\ref{fig:notetaking}, the \strategy{} effectively simulates how humans take notes during long and complex conversations, facilitating the retention of key details and maintaining focus. As shown in Table~\ref{tab:exp_improve}, the improvement is observed in the long-term memory ability, with MR increasing by an average of +0.97. Furthermore, the note-taking mechanism enhances the model's information extraction capability by +0.88, and improves information update by +0.84. Moreover, the structured clarity provided by the notes allows the model to concentrate more effectively on relevant details within the conversation, resulting in a +0.52 improvement in answer refusal.

\begin{table}[!t]
    \centering
    \small
\resizebox{0.47\textwidth}{!}{
\begin{tabular}{l|cccccc}
\toprule
\textbf{Models} & IE & IU & MR & AR  \\
\midrule
LLaVA-1.5-7B & 0.91 & 0.31 & 0.22 & 0.28 \\
~~~~+ \strategy{} & \textbf{2.57} & \textbf{2.06} & \textbf{2.36} & \textbf{0.46}\vspace{-4pt} \\
\scriptsize\color[RGB]{0,120,0}& \scriptsize\color[RGB]{0,120,0} (+1.66) &\scriptsize\color[RGB]{0,120,0} (+1.75) & \scriptsize\color[RGB]{0,120,0} (+2.14) &\scriptsize\color[RGB]{0,120,0} (+0.18) \\
\midrule
MiniCPM-8B & 4.08 & 2.98 & 3.65 & 3.78 \\
~~~~+ \strategy{} & \textbf{4.23} & \textbf{3.76} & \textbf{4.02} & \textbf{3.92} \vspace{-4pt} \\
\scriptsize\color[RGB]{0,120,0}& \scriptsize\color[RGB]{0,120,0} (+0.15) &\scriptsize\color[RGB]{0,120,0} (+0.78) & \scriptsize\color[RGB]{0,120,0} (+0.37) & \scriptsize\color[RGB]{0,120,0} (+0.14) \\
\midrule
QwenVL-2B & 2.16 & 1.41 & 1.93 & 2.33  \\
~~~~+ \strategy{} & \textbf{3.71} & \textbf{2.43} & \textbf{3.40} & \textbf{2.84} \vspace{-4pt} \\
\scriptsize\color[RGB]{0,120,0}& \scriptsize\color[RGB]{0,120,0} (+1.55) &\scriptsize\color[RGB]{0,120,0} (+0.82) & \scriptsize\color[RGB]{0,120,0} (+1.47) & \scriptsize\color[RGB]{0,120,0} (+0.51) \\
\midrule
LLaVA-Next-0.5B & 2.32 & 1.99 & 2.67 & 1.12 \\
~~~~+ \strategy{} & \textbf{3.84} & \textbf{3.04} & \textbf{3.88} & \textbf{2.03} \vspace{-4pt} \\
\scriptsize\color[RGB]{0,120,0}& \scriptsize\color[RGB]{0,120,0} (+1.52) &\scriptsize\color[RGB]{0,120,0} (+1.05) & \scriptsize\color[RGB]{0,120,0} (+1.21) & \scriptsize\color[RGB]{0,120,0} (+0.91) \\
\midrule
LLaVA-OneVision-72B & 4.06 & 4.01 & 4.17 & 2.52 \\
~~~~+ \strategy{} & \textbf{4.28} & \textbf{4.31} & \textbf{4.38} & \textbf{3.46} \vspace{-4pt} \\
\scriptsize\color[RGB]{0,120,0}& \scriptsize\color[RGB]{0,120,0} (+0.22) &\scriptsize\color[RGB]{0,120,0} (+0.30) & \scriptsize\color[RGB]{0,120,0} (+0.21) & \scriptsize\color[RGB]{0,120,0} (+0.94) \\
\midrule
GPT-4o & 4.35 & 4.28 & 4.31 & 3.06 \\
~~~~+ \strategy{} & \textbf{4.51} & \textbf{4.62} & \textbf{4.73} & \textbf{3.51} \vspace{-4pt} \\
\scriptsize\color[RGB]{0,120,0}& \scriptsize\color[RGB]{0,120,0} (+0.16) &\scriptsize\color[RGB]{0,120,0} (+0.34) & \scriptsize\color[RGB]{0,120,0} (+0.42) & \scriptsize\color[RGB]{0,120,0} (+0.45) \\
\bottomrule
    \end{tabular}}
    \vspace{-0.2cm}
    \caption{Performance of \strategy{} across four conversational core abilities in \ours{}.}
    \label{tab:exp_improve}
    \vspace{-0.3cm}
\end{table}

\vspace{-0.2cm}
\section{Conclusions \& Limitations}
\vspace{-0.1cm}

In this paper, we introduce \ours{}, the first multi-image open-ended conversation benchmark to evaluate the six conversation abilities of MLLMs. Our comprehensive analysis identifies four common failure patterns: long-term memory degradation, inadequate updating of factual knowledge, accumulated assumption of error propagation, and reluctance to “say no.” To mitigate these, we propose the \strategy{} strategy, which stores key user preferences and facts by using structured prompts.

\vspace{0.1cm}
\noindent\textbf{Limitations:} We clarify the limitations: \textit{(i):} While \ours{} covers multiple domains, it may not encompass all real-world dialogue types (\textit{e.g.,} population distribution and languages) and requires further exploration. \textit{(ii):} Although the \strategy{} improves model performance, the note generation process can be computationally intensive.

\bibliography{custom}

\appendix
\newpage
\onecolumn

\section{Open-ended Conversations}

Open-ended conversation is a flexible and unconstrained form of conversations~\citep{fernyhough1996dialogic,barnden2014open}, allowing users to engage freely without predefined limits~\citep{xiao2020tell,yoshimura2024synlogue}. In these conversations, the user has full control, enabling them to express thoughts and emotions openly~\citep{elfenbein2022we,seo2024chacha}, as well as delve into topics in greater depth to uncover insights or solutions~\citep{aziza2021teacher,sun2022relationship}. This type of conversation is the most common mode of interaction for general users in MLLM chat systems~\citep{fu2024mme,liu2024mibench}. Recent studies have shown that LLMs can exhibit reasoning capabilities akin to human-like problem-solving~\citep{jin2024impact,jin2025massive}. Moreover, in computer vision~\citep{tang2024hunting,tang2023duat,tang2024neighbor,zhao2024sfc,wang2022stepwise,xu2024toward,xu2024polyp,hu2025ophnet,hu2024ophclip,chen2024meta,xiong2024sam2,hu2024diffusion,trinh2024sight,tang2024discriminating}, open-ended conversations facilitate more natural human-AI interactions, enabling models to provide richer visual explanations, refine image generation based on iterative feedback, and support interactive learning for complex vision tasks.

\section{Ethics Statements}

The widespread availability of data and powerful analytical tools play a pivotal role in research~\citep{pfenninger2017importance,super1982relative}, but also come with the risk of misuse~\citep{pasquetto2024research,shimron2022implicit}. Therefore, ethical oversight is crucial, it involves issues of privacy~\citep{pina2024data,lacroix2019big}, ownership~\citep{andrews2023ethical,kapusta2024protecting}, consent~\citep{mckeown2021ethical,longpre2024consent}, and purpose of use~\citep{paullada2021data,padmapriya2024ethical}. Based on the definitions and related issues concerning dataset ethics, we make the following statement about \ours{}:

The data collection for our study is conducted with the informed consent of all participants, ensuring their privacy and autonomy are fully protected. All participants are fully aware of and voluntarily engage in the annotation process. We implement a rigorous review mechanism to ensure the data is free from personally identifiable information, offensive material, and violent content. However, given the limitations of manual inspection, some residual information may still remain, and completely eliminating such content remains a challenging task. Furthermore, since the data originates from real conversations, it may contain a small amount of inadvertently misleading information, which could impact the model’s performance on benchmark tests. We release this data exclusively for research purposes, allowing researchers to explore the performance of multimodal large language models (MLLMs) in real-world dialogue contexts. However, researchers must approach this dataset with utmost caution and ethical consideration. Our goal is to contribute to the accumulation of knowledge while ensuring that our research findings are applied ethically. In the future, we will continue to release updated versions of the dataset, expanding both its volume and comprehensiveness, while further filtering out offensive content and misinformation.

\section{\platform}
\label{Dialogflow}
 
Collecting data through online platforms is an efficient method to gather large volumes of valuable data~\citep{panduman2024survey,wang2024occupant}. It offers a cost-effective and scalable solution for data collection, adapting to varying needs over time~\citep{pamucar2024selection,langer2024claid}. Therefore, we establish {\platform} to collect user conversation data.

\begin{figure}
    \centering
    \includegraphics[width=0.55\linewidth]{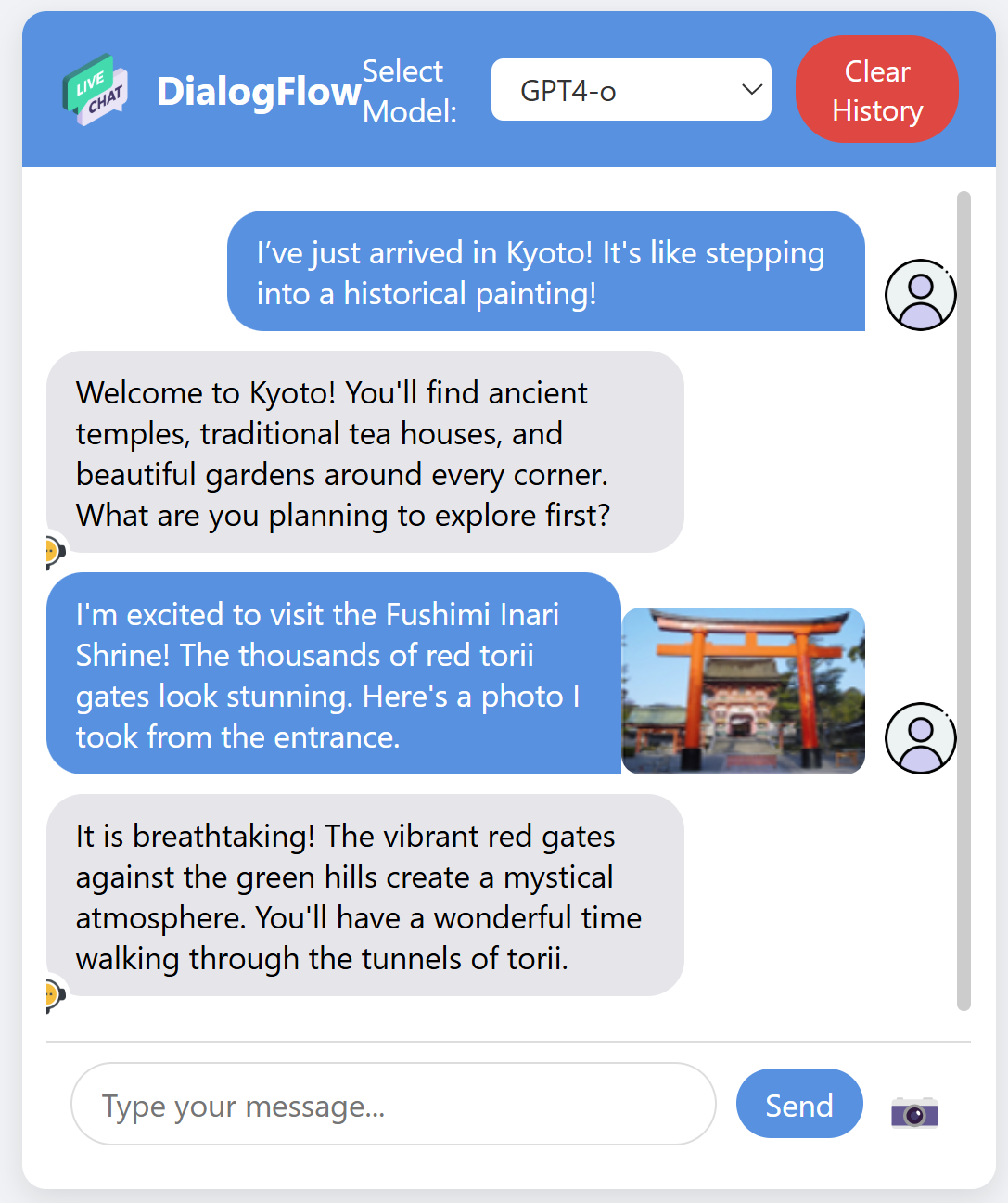}
    \caption{The page of \platform }
    \label{fig:flatform}
\end{figure}

Due to GPU source limitations, {\platform} is not open to the public and is only undergoing internal testing within the campus. The interface of {\platform} is shown in Fig.\ref{fig:flatform}. Users can add image inputs by clicking the camera icon at the bottom right corner and switch between different models by changing the tabs at the top. The platform is free to use since its launch on April 6, 2024. We use dozens of A100 GPUs to host our website and run open-source models for user conversations, while closed-source models interact via APIs. All the models available on the platform are listed in Table~\ref{tab:platform}. {\platform} has a total of 354 users, with ages ranging from 18 to 47. The majority of them are university students and staff. Users are required to accept the Terms of Use to consent to the public release of their conversation data.

\begin{table}[!t]
\centering
\footnotesize
\vspace{-1em}
\resizebox{\textwidth}{!}{
\begin{tabular}{@{} lcccc @{}}
\toprule
Model  & Size & Vision Encoder & LLM \\
\midrule
\midrule
LLaVA-V1.5~\citep{liu2023improvedLLaVA}  & 7B,13B & CLIP ViT-L/14~\citep{zhang2022clip} & Vicuna-v1.5~\citep{zheng2023judging}\\
MiniCPM~\citep{yao2024minicpm}  &8B & SigLIP-SoViT-400m/14~\citep{zhai2023sigmoid} & Llama3-Instruct~\citep{dubey2024llama}\\
LLaVA-Next~\citep{li2024llava} &0.5B,7B & SigLIP-400M~\citep{zhai2023sigmoid} & Qwen1.5~\citep{bai2023qwen} \\
Qwen2VL~\citep{wang2024qwen2} &2B,7B,72B & ViT trained from scratch~\citep{alexey2020image} & Qwen2~\citep{qwen2} \\
LLaVA-OneVision~\citep{li2024llava} &0.5B,7B & SigLIP-400M~\citep{zhai2023sigmoid} & Qwen2~\citep{qwen2}\\
VILA1.5~\citep{lin2024vila}  &3B,8B & SigLIP-400M~\citep{zhai2023sigmoid} &  Llama3~\citep{grattafiori2024llama}\\
mplug-Ow3~\citep{ye2024mplug} &1B,2B,7B &  SigLIP-400M~\citep{zhai2023sigmoid} & Qwen2~\citep{qwen2}\\
\midrule
GPT-4o~\citep{islam2024gpt}  &~300B &-&-\\
Claude-3.5-sonnet~\citep{yao2024minicpm} &175B &-&-\\
Gemini-1.5 Pro~\citep{team2024gemini}  & 175B &-&-\\
DeepSeek-V3~\citep{liu2024deepseek} &671B &-&-\\
\midrule
\bottomrule
\end{tabular}}
\caption{MLLMs in \platform.}
\vspace{-1em}
\label{tab:platform}
\end{table}

\section{Examples of \ours{}}
We list 5 samples of \ours{}, they are conversations about travel (Fig.~\ref{fig:travel}), dancing (Fig.~\ref{fig:dance}), physics (Fig.~\ref{fig:physic}), water parks (Fig.~\ref{fig:waterpark}), and dessert making (Fig.~\ref{fig:dessert}). The domains of these conversations are comprehensive, natural, and closely aligned with real-world usage scenarios. Additionally, the evaluation questions for the conversations are manually annotated and checked by our team. All of this demonstrates the high quality and real-world relevance of our data.

\section{Topic Classification Network}
\label{Topic net}

The 14 predefined categories are carefully chosen to ensure comprehensive coverage of real-world conversational topics, and their selection is grounded in the characteristics of our dataset and relevant studies on human conversations~\citep{kim2018dialog,zhao2019review,zhang2020recent,algherairy2024review}.

To train the topic classification network, we manually annotate a portion of the data, label the date with 14 predefined categories, each containing 50 dialogue samples. We split the data into 90\% for training and 10\% for testing. For the classification, we use the all-mpnet-base-v2 model from SentenceTransformers~\citep{piao2021scholarly} to generate text embeddings. These embeddings are passed through a two-layer MLP for classification. The model ultimately achieve over 95\% accuracy on the test set.

\section{Statics of conversation domains in \ours{}}
\label{Statics of conversation domains}
\begin{figure*}[p]
 \centering
  \captionsetup{justification=raggedright, singlelinecheck=false}
  \includegraphics[width=0.99\textwidth]{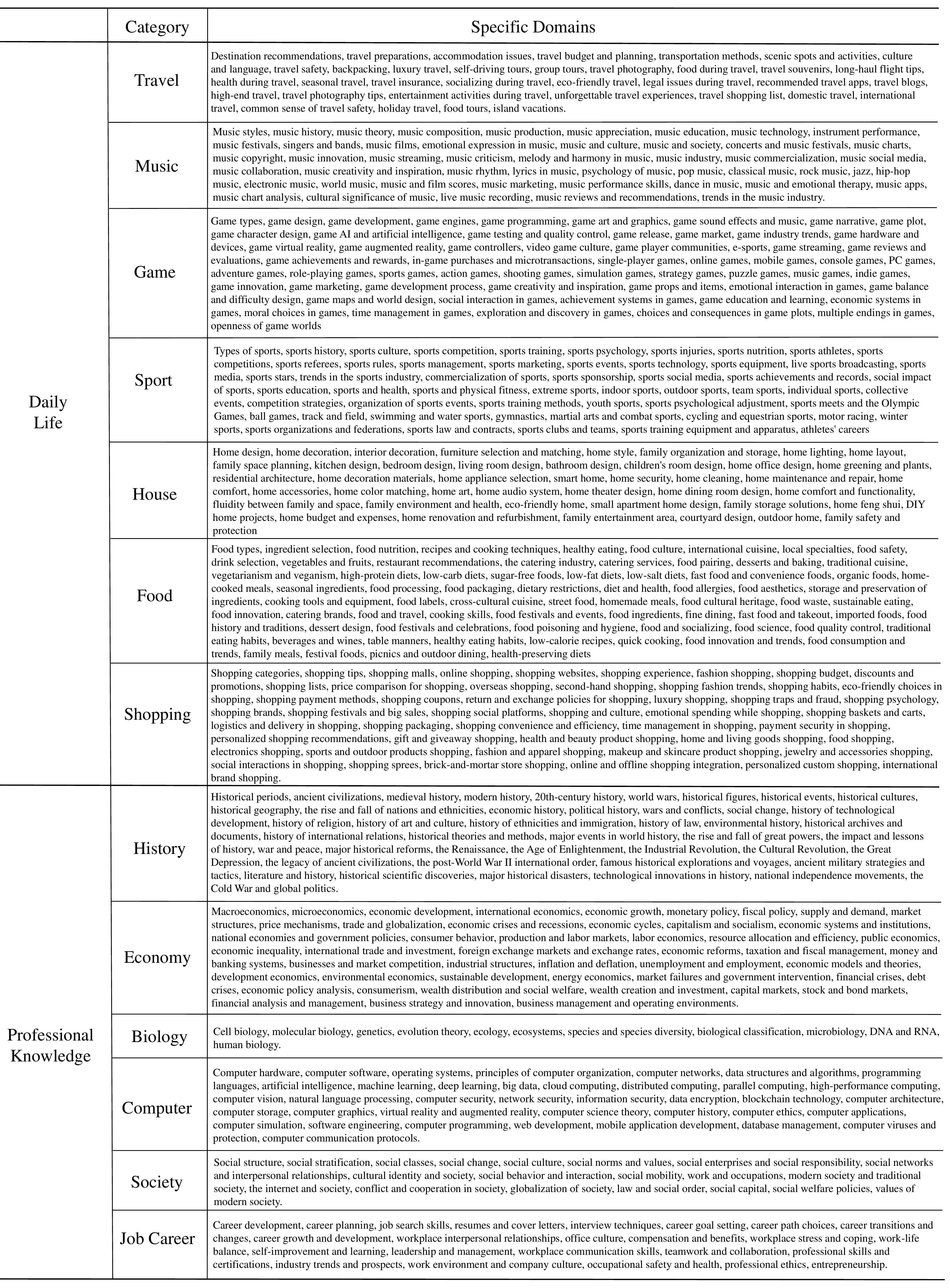} 
  \vspace{-0.2cm}
  \centering
  \caption{List of partial existing domains of conversations in \ours{}.}
  \vspace{-0.3cm}
  \label{fig:category}
\end{figure*}

Although we have categorized the existing conversation data into 14 categories for convenience in statistical analysis, the actual diversity of the data far exceeds this classification. If we increase the granularity of the categories, our data would encompass more domains. According to our statistics, we have data from 874 different domains. Due to space limitations, we select a subset of domains, as shown in Fig.~\ref{fig:category}, conversation data in \ours{} is comprehensive and diverse, covering various aspects of real-world scenarios. Therefore, the evaluation of \ours{} provides valuable insights of MLLMs' performance in practical open-ending conversations.

\section{Experimental Settings}

LLaVA-V1.5-7B, LLaVA-V1.5-13B, MiniCPM-8B, LLaVA-Next-0.5B, LLaVA-Next-7B, Qwen2VL-2B, Qwen2VL-7B, LLaVA-OneVision-0.5B, LLaVA-OneVision-7B, VILA1.5-3B, VILA1.5-8B, mplug-Ow3-1B, mplug-Ow3-2B, mplug-Ow3-7B use 16-bit floating-point precision, while Qwen2VL-72B and LLaVA-OneVision-72B use 4-bit quantization. Their output length is limited to a maximum of 256 tokens. The models utilize the default values for Temperature, Top-k Sampling, and Top-p Sampling as specified in their Hugging Face inference code. 

\section{Why DeepSeek-V3 is Classified as Close-Sourced in Experiment}
Although the DeepSeek-V3 has been open-sourced on Hugging Face, its large parameter size makes local inference very resource-consuming. Therefore, we use the API for testing. In the experiment, we categorized it under the close-sourced section, as its parameter size is similar to that of close-sourced models, both exceeding 175B parameters.

\section{Prompt Template and Human Criteria for Evaluation}
\label{Prompt templete}

In recent years, an increasing number of benchmark studies have utilized LLM as an evaluation tool, due to its accuracy, logical reasoning, and ability to follow instructions~\citep{chenmllm,li2024generation,gu2024survey,ye2024justice}. Building on these advancements, \ours{} also applies GPT-4o for scoring the model's information extraction, cross-turn reasoning, information update, image management, long-term memory recall, and answer refusal capabilities. The detailed evaluation prompts corresponding to each capability are shown in Fig.~\ref{fig:IE_prompt}, Fig.~\ref{fig:CR_prompt}, Fig.~\ref{fig:IU_prompt}, Fig.~\ref{fig:IM_prompt}, and Fig.~\ref{fig:MR_prompt}. We also conducte manual evaluation of the capabilities in cross-turn reasoning, information update, and memory recall. The three evaluation criteria are as follows:

\noindent\textbf{(1) Cross-turn reasoning (CR):} 

\noindent\textbf{Goal:} To assess how well the model integrates and utilizes information across multiple dialogue turns to answer complex questions.

\noindent\textbf{Criteria:} \textit{Full reasoning (5 points): }the model accurately integrates information from multiple dialogue turns and provides a clear answer that reflects all key details, demonstrating full understanding of the context; \textit{Partial reasoning (3-4 points): }the model integrates some of the conversation’s information, but the answer does not fully reflect all necessary context, it is partially correct; \textit{Error in Reasoning (2 points): }the model fails to integrate information properly, providing an unclear or erroneous reasoning chain, leading to an inaccurate answer; \textit{Lack of Reasoning (1 point): }the model does not demonstrate effective reasoning, providing a vague or irrelevant answer without appropriately incorporating the conversation’s context.

\noindent\textbf{(2) Information Update (IU):}

\noindent\textbf{Goal:} To evaluate how well the model tracks and updates factual information provided by the user during the conversation.

\noindent\textbf{Criteria:} \textit{Complete Update (5 points):} the model accurately and promptly updates new factual information, incorporating this into later responses and maintaining consistency throughout the conversation; \textit{Partial Update (3-4 points):} the model updates some information but fails to fully incorporate the changes in subsequent responses. There may be some omissions or incomplete updates; \textit{Failure to Update (2 points):} the model does not recognize the change in factual information and continues to rely on outdated data, leading to inconsistent responses; \textit{Incorrect Update (1 point):} the model incorrectly updates information, causing contradictions with prior details or incorrect facts to be incorporated in later responses.

\noindent\textbf{(3) Memory Recall (MR):}

\noindent\textbf{Goal:} To assess how well the model maintains long-term memory of the conversation and recalls relevant details in later dialogue turns.

\noindent\textbf{Criteria:} \textit{Accurate Recall (5 points):} the model accurately recalls key information (e.g., user preferences, previous conversation details) from earlier dialogue turns and integrates this information effectively in later responses; \textit{Partial Recall (3-4 points):} the model recalls some details but misses out on others, leading to slightly disjointed or incomplete answers; \textit{Memory Loss (2 points):} the model forgets important details over the course of the conversation, leading to inconsistent or disconnected responses that lack coherence; \textit{Incorrect Recall (1 point):} the model recalls information incorrectly (e.g., mixing up user preferences or repeating outdated facts), leading to answers that are inaccurate or misleading.

\section{Attention Calculation}
\label{Attention Calculation}

The phenomenon of attention imbalance distribution in models during dialogue shown in Fig~\ref{fig:attention}, is consistent with recent research~\citep{jawale2024human}. The following are the detailed steps for calculating attention: we randomly selecte 100 conversations with 15 turns to visualize the attention. For these dialogues, we perform manual special processing, where we label the information locations related to subsequent memory questions (\textit{i.e.,}IE, IM, MR) as the 'golden position', referred as $G$. For example, if answering an IE question requires extracting information from turn 2 and turn 4, the golden position for this IE question would be turns 2 and 4. The role of the golden label in subsequent calculations will be explained in detail. For the attention calculation model, we selected LLaVA-Next-0.5B, LLaVA-OneVision-72B, MiniCPM-8B, and QwenVL-2B. These choices aim to cover a broad range of model families and parameter sizes.

When the model is tasked with answering memory-related questions, the input typically consists of a dialogue history and an memory evaluation question. Let the input be \texttt{query} = $[x_1, x_2, \ldots, x_n, q_s]$, where $[x_1, x_2, \ldots, x_n]$ represents the conversation history of the model, and each turn $x_i$ (where $1 \leq i \leq n$) may contain image tokens and text tokens, and $q_s$ represents the current evaluation question. To compute the self-attention weights for each turn, we define a function \texttt{Attn}: $\mathcal{X} \times \mathcal{N} \to \mathbb{R}$, which computes the average attention weight assigned to each turn $x_i$, where each turn $x_i = \{x_{i,j}\}^{N_i}_{j=1}$ contains $N_i$ tokens. Only the turns in the golden position are considered in the attention calculation. The other turns are excluded to avoid interference from irrelevant information. We compute the average attention weight for each golden position turn using the following formula:

\[
\text{Attn}(q_s, i) = \frac{1}{N_i} \sum_{j=1}^{N_i} \text{attn}(x_{i,j}),\quad x_i \in G
\]

\section{Details on \strategy{}}
\label{Note taking}

The \strategy{} employs another MLLM for note-taking, with model selection including open-source models like LLaVA-One-Vision-7B, MiniCPM-8B, as well as closed-source models like GPT-4o. Since the task of note-taking is easier than conversation, the choice of different models has little impact on note-taking performance. The algorithm for our proposed \strategy{} strategy is in Alg.~\ref{alg:ours}. Due to the limitations of space in the main text, we have only presented a portion of the experimental results. Table~\ref{tab:exp_improve_2} shows the complete experimental results.

\begin{algorithm}
\caption{\strategy{} algorithm}
\label{alg:ours}
\begin{algorithmic}[1] %
\Require MLLM A (notetaker), MLLM B (test model), user input $\mathcal{T}$, evaluation questions $\mathcal{Q}$. 

\noindent\textit{Step1: Takes Notes}
\State $\textit{Note} = \{\}$
\For{$\mathcal{T}_i$ \textbf{in} $T$}
    \State $history = \text{append}(T_i)$ 
    \State $\mathcal{R}_i = Chat(\text{B}, history)$ %
    
    \State $\textit{Note} = \textit{Take\_Note} (\text{A}, T_i,\textit{Note})$
    \State $history = \text{append}(R_i)$ 
\EndFor

\noindent\textit{Step 2: Help to response}
\For{$\mathcal{Q}_i$ \textbf{in} $Q$}
    \State $history = \text{append}(Q_i)$ 
    \State $answer = \text{Chat}(\text{B}, (history, \textit{Note}))$
    \State \textbf{output} $response_i$
\EndFor
\end{algorithmic}
\end{algorithm}

 \begin{figure}[H]  
  \centering
  \includegraphics[width=\linewidth]{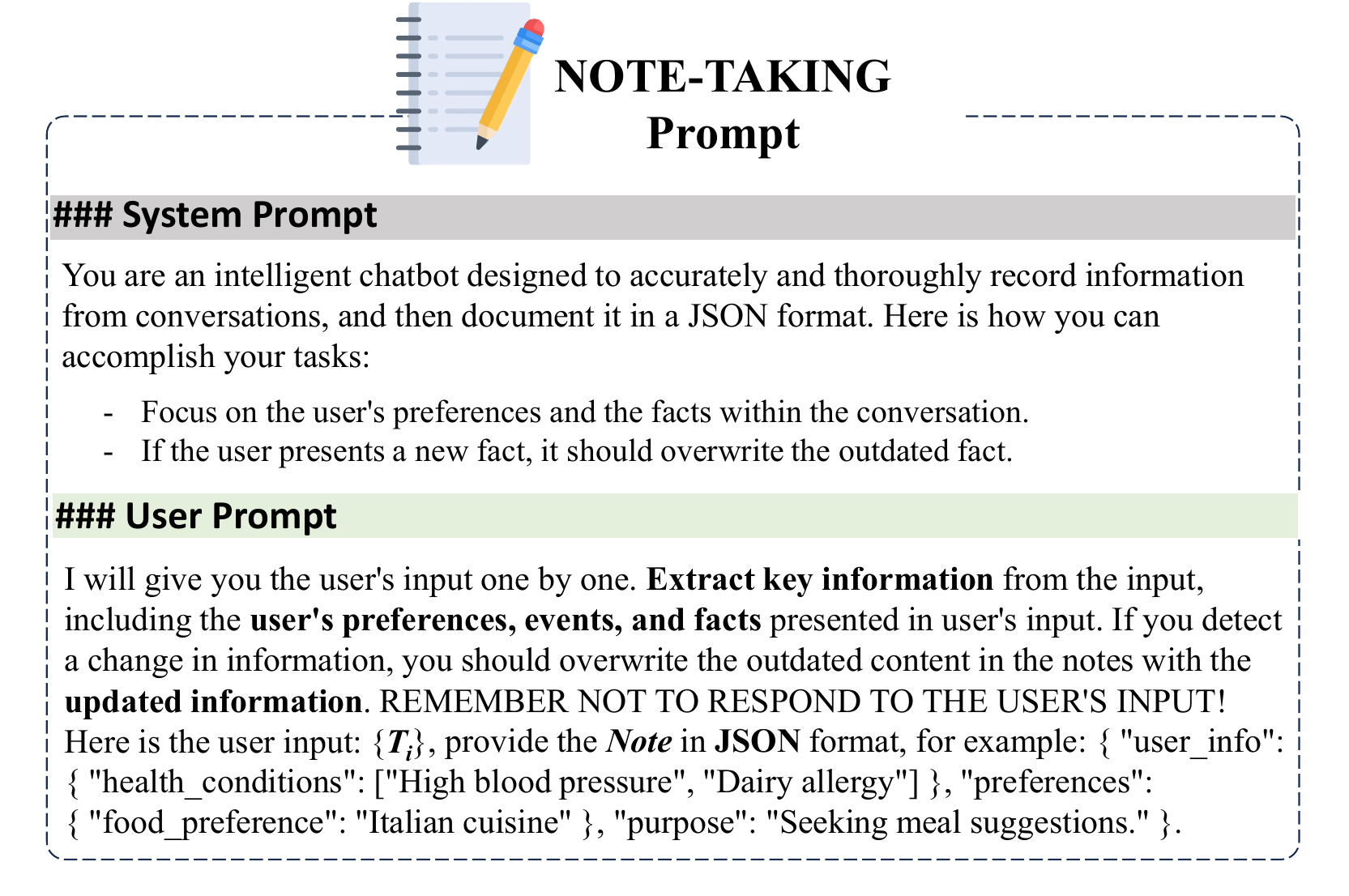} 
  \caption{The prompt template of \textit{Take\_Note} function in Alg.~\ref{alg:ours}.}

  \label{fig:CR_prompt}

\end{figure}

\begin{table}[!h]
    \centering
    \small
\resizebox{0.8\textwidth}{!}{
\begin{tabular}{l|cccccccc}
\toprule
\textbf{Models} & IE & IU & MR & AR & CR & IM \\
\midrule
LLaVA-1.5-7B & 0.91 & 0.31 & 0.22 & 0.28 & 1.08 & 0.52 \\
~~~~+ \strategy{}(GPT-4o) & \textbf{2.57} & \textbf{2.06} & \textbf{2.36} & \textbf{0.46} & \textbf{1.22} & \textbf{0.54}
 \vspace{-4pt} \\
\scriptsize\color[RGB]{0,120,0}& \scriptsize\color[RGB]{0,120,0} (+1.66) &\scriptsize\color[RGB]{0,120,0} (+1.75) & \scriptsize\color[RGB]{0,120,0} (+2.14) &\scriptsize\color[RGB]{0,120,0} (+0.18) &\scriptsize\color[RGB]{0,120,0} (+0.14) &\scriptsize\color[RGB]{0,120,0} (+0.02) \\
~~~~+ \strategy{}(MiniCPM) & \textbf{2.33} & \textbf{1.78} & \textbf{2.11} & \textbf{0.35} & \textbf{1.19} & \textbf{0.56}
 \vspace{-4pt} \\
\scriptsize\color[RGB]{0,120,0}& \scriptsize\color[RGB]{0,120,0} (+1.42) &\scriptsize\color[RGB]{0,120,0} (+1.47) & \scriptsize\color[RGB]{0,120,0} (+1.89) &\scriptsize\color[RGB]{0,120,0} (+0.07) &\scriptsize\color[RGB]{0,120,0} (+0.11) &\scriptsize\color[RGB]{0,120,0} (+0.04) \\
\midrule
MiniCPM-8B & 4.08 & 2.98 & 3.65 & 3.78 & 3.94 & 3.47 \\
~~~~+ \strategy{}(GPT-4o)  & \textbf{4.23} & \textbf{3.76} & \textbf{4.02} & \textbf{3.92}  & \textbf{4.13}  & \textbf{3.50} \vspace{-4pt} \\
\scriptsize\color[RGB]{0,120,0}& \scriptsize\color[RGB]{0,120,0} (+0.15) &\scriptsize\color[RGB]{0,120,0} (+0.78) & \scriptsize\color[RGB]{0,120,0} (+0.37) & \scriptsize\color[RGB]{0,120,0} (+0.14) &\scriptsize\color[RGB]{0,120,0} (+0.09) &\scriptsize\color[RGB]{0,120,0} (+0.03) \\
~~~~+ \strategy{}(MiniCPM) & \textbf{4.12} & \textbf{3.45} & \textbf{3.74} & \textbf{3.81} & \textbf{4.02} & \textbf{3.52}
 \vspace{-4pt} \\
\scriptsize\color[RGB]{0,120,0}& \scriptsize\color[RGB]{0,120,0} (+0.04) &\scriptsize\color[RGB]{0,120,0} (+0.47) & \scriptsize\color[RGB]{0,120,0} (+0.09) &\scriptsize\color[RGB]{0,120,0} (+0.03) &\scriptsize\color[RGB]{0,120,0} (+0.08) &\scriptsize\color[RGB]{0,120,0} (+0.05) \\
\midrule
QwenVL-2B & 2.16 & 1.41 & 1.93 & 2.33 & 2.85 & 1.27 \\
~~~~+ \strategy{}(GPT-4o)  & \textbf{3.71} & \textbf{2.43} & \textbf{3.40} & \textbf{2.84}  & \textbf{3.47}  & \textbf{1.36} \vspace{-4pt} \\
\scriptsize\color[RGB]{0,120,0}& \scriptsize\color[RGB]{0,120,0} (+1.55) &\scriptsize\color[RGB]{0,120,0} (+0.82) & \scriptsize\color[RGB]{0,120,0} (+1.47) & \scriptsize\color[RGB]{0,120,0} (+0.51) &\scriptsize\color[RGB]{0,120,0} (+0.62) &\scriptsize\color[RGB]{0,120,0} (+0.09) \\
~~~~+ \strategy{}(MiniCPM) & \textbf{3.22} & \textbf{2.01} & \textbf{3.21} & \textbf{2.58} & \textbf{3.26} & \textbf{1.32}
 \vspace{-4pt} \\
\scriptsize\color[RGB]{0,120,0}& \scriptsize\color[RGB]{0,120,0} (+1.06) &\scriptsize\color[RGB]{0,120,0} (+0.60) & \scriptsize\color[RGB]{0,120,0} (+1.28) &\scriptsize\color[RGB]{0,120,0} (+0.25) &\scriptsize\color[RGB]{0,120,0} (+0.41) &\scriptsize\color[RGB]{0,120,0} (+0.05) \\
\midrule
LLaVA-Next-0.5B & 2.32 & 1.99 & 2.67 & 1.12 & 2.89 & 1.87 \\
~~~~+ \strategy{}(GPT-4o)  & \textbf{3.84} & \textbf{3.04} & \textbf{3.88} & \textbf{2.03}  & \textbf{3.62}  & \textbf{2.07} \vspace{-4pt} \\
\scriptsize\color[RGB]{0,120,0}& \scriptsize\color[RGB]{0,120,0} (+1.52) &\scriptsize\color[RGB]{0,120,0} (+1.05) & \scriptsize\color[RGB]{0,120,0} (+1.21) & \scriptsize\color[RGB]{0,120,0} (+0.91) &\scriptsize\color[RGB]{0,120,0} (+0.73) &\scriptsize\color[RGB]{0,120,0} (+0.20) \\
~~~~+ \strategy{}(MiniCPM) & \textbf{3.68} & \textbf{2.77} & \textbf{3.59} & \textbf{2.06} & \textbf{3.38} & \textbf{1.94}
 \vspace{-4pt} \\
\scriptsize\color[RGB]{0,120,0}& \scriptsize\color[RGB]{0,120,0} (+1.36) &\scriptsize\color[RGB]{0,120,0} (+0.78) & \scriptsize\color[RGB]{0,120,0} (+0.92) &\scriptsize\color[RGB]{0,120,0} (+0.94) &\scriptsize\color[RGB]{0,120,0} (+0.49) &\scriptsize\color[RGB]{0,120,0} (+0.07) \\
\midrule
LLaVA-OneVision-72B & 4.06 & 4.01 & 4.17 & 2.52 & 4.08 & 3.24 \\
~~~~+ \strategy{}(GPT-4o)  & \textbf{4.28} & \textbf{4.31} & \textbf{4.38} & \textbf{3.46}  & \textbf{4.16}  & \textbf{3.31} \vspace{-4pt} \\
\scriptsize\color[RGB]{0,120,0}& \scriptsize\color[RGB]{0,120,0} (+0.22) &\scriptsize\color[RGB]{0,120,0} (+0.30) & \scriptsize\color[RGB]{0,120,0} (+0.21) & \scriptsize\color[RGB]{0,120,0} (+0.94) &\scriptsize\color[RGB]{0,120,0} (+0.08) &\scriptsize\color[RGB]{0,120,0} (+0.07)\\
~~~~+ \strategy{}(MiniCPM) & \textbf{4.13} & \textbf{4.28} & \textbf{4.26} & \textbf{3.20} & \textbf{4.13} & \textbf{3.26}
 \vspace{-4pt} \\
\scriptsize\color[RGB]{0,120,0}& \scriptsize\color[RGB]{0,120,0} (+0.07) &\scriptsize\color[RGB]{0,120,0} (+0.27) & \scriptsize\color[RGB]{0,120,0} (+0.09) &\scriptsize\color[RGB]{0,120,0} (+0.68) &\scriptsize\color[RGB]{0,120,0} (+0.05) &\scriptsize\color[RGB]{0,120,0} (+0.02) \\
\midrule
GPT-4o & 4.35 & 4.28 & 4.31 & 3.06 & 4.38 & 4.12 \\
~~~~+ \strategy{}(GPT-4o)  & \textbf{4.51} & \textbf{4.62} & \textbf{4.73} & \textbf{3.51}  & \textbf{4.41}  & \textbf{4.18} \vspace{-4pt} \\
\scriptsize\color[RGB]{0,120,0}& \scriptsize\color[RGB]{0,120,0} (+0.16) &\scriptsize\color[RGB]{0,120,0} (+0.34) & \scriptsize\color[RGB]{0,120,0} (+0.42) & \scriptsize\color[RGB]{0,120,0} (+0.45) &\scriptsize\color[RGB]{0,120,0} (+0.03) &\scriptsize\color[RGB]{0,120,0} (+0.06) \\
~~~~+ \strategy{}(MiniCPM) & \textbf{4.46} & \textbf{4.45} & \textbf{4.37} & \textbf{3.32} & \textbf{4.40} & \textbf{4.16}
 \vspace{-4pt} \\
\scriptsize\color[RGB]{0,120,0}& \scriptsize\color[RGB]{0,120,0} (+0.11) &\scriptsize\color[RGB]{0,120,0} (+0.17) & \scriptsize\color[RGB]{0,120,0} (+0.06) &\scriptsize\color[RGB]{0,120,0} (+0.26) &\scriptsize\color[RGB]{0,120,0} (+0.02) &\scriptsize\color[RGB]{0,120,0} (+0.04) \\
\bottomrule
    \end{tabular}}
    \vspace{-0.2cm}
    \caption{Performance of \strategy{} across all conversational core abilities in \ours{}.}
    \label{tab:exp_improve_2}
    \vspace{-0.3cm}
\end{table}

\begin{figure*}[p]
 \centering
  \captionsetup{justification=raggedright, singlelinecheck=false}
  \includegraphics[width=\textwidth]{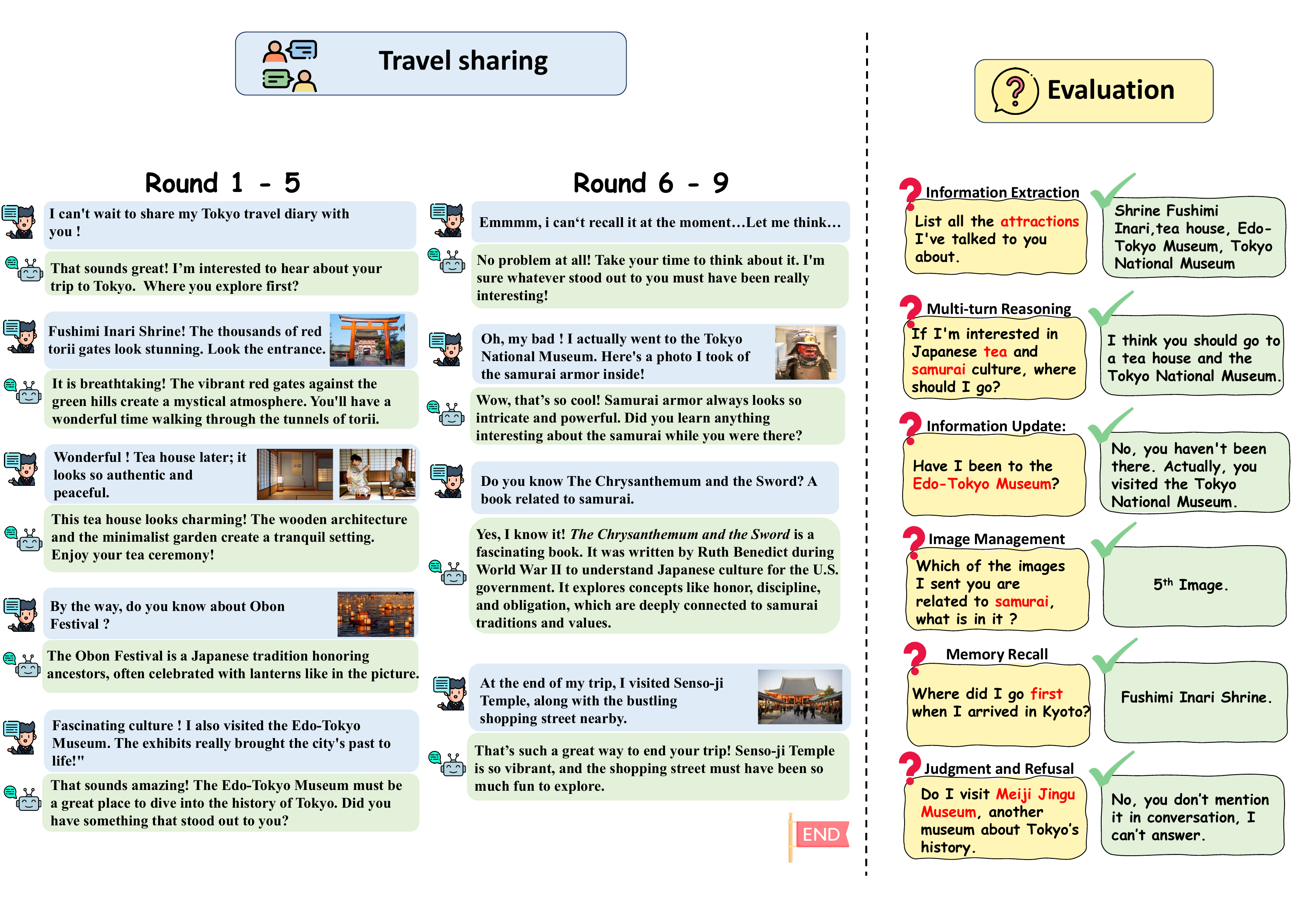} 
  \vspace{-0.2cm}
  \caption{A sample from \ours{}, the dialog domain is travel sharing, consisting of 9 conversation turns and 6 image inputs.}
  \label{fig:travel}
\end{figure*}

\begin{figure*}[p]
 \centering
  \captionsetup{justification=raggedright, singlelinecheck=false}
  \includegraphics[width=\textwidth]{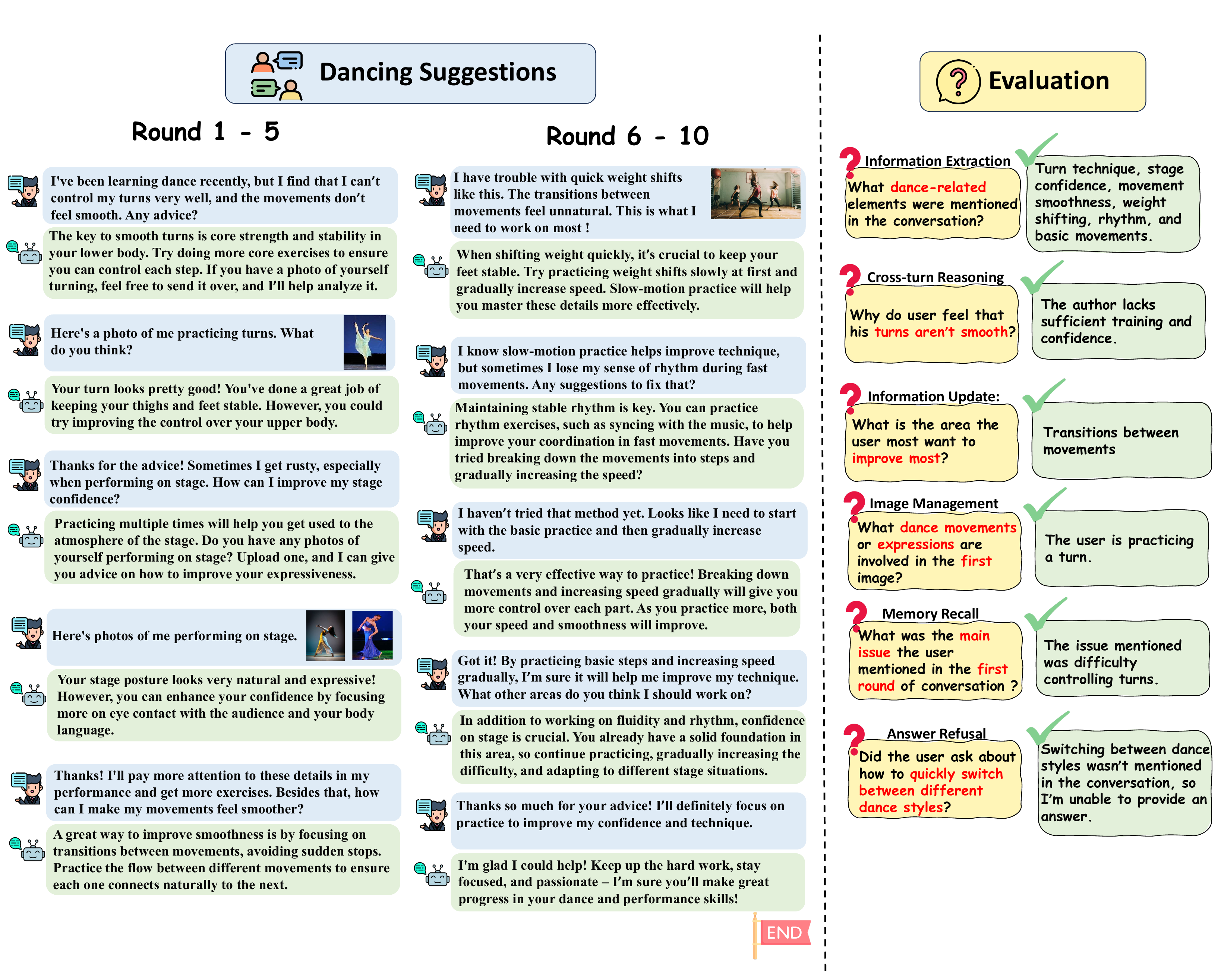} 
  \vspace{-0.2cm}
  \caption{A sample from \ours{}, the dialogue domain is dancing suggestions, consisting of 10 conversation turns and 4 image inputs.}
  \label{fig:dance}
\end{figure*}

\begin{figure*}[p]
 \centering
  \captionsetup{justification=raggedright, singlelinecheck=false}
  \includegraphics[width=\textwidth]{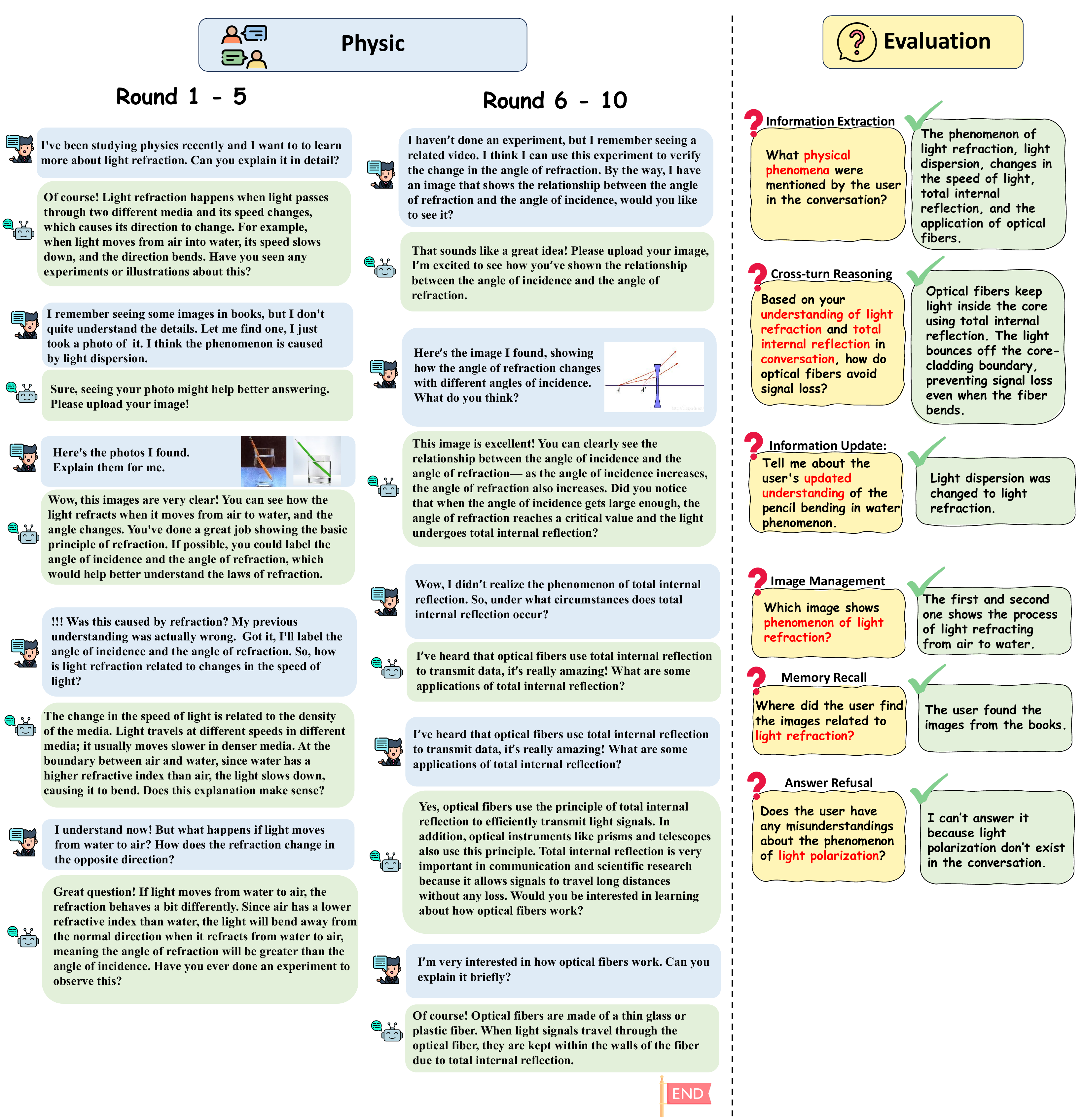} 
  \vspace{-0.2cm}
  \caption{A sample from \ours{}, the dialogue domain is physic, consisting of 10 conversation turns and 3 image inputs.}
  \label{fig:physic}
\end{figure*}

\begin{figure*}[p]
 \centering
  \captionsetup{justification=raggedright, singlelinecheck=false}
  \includegraphics[width=\textwidth]{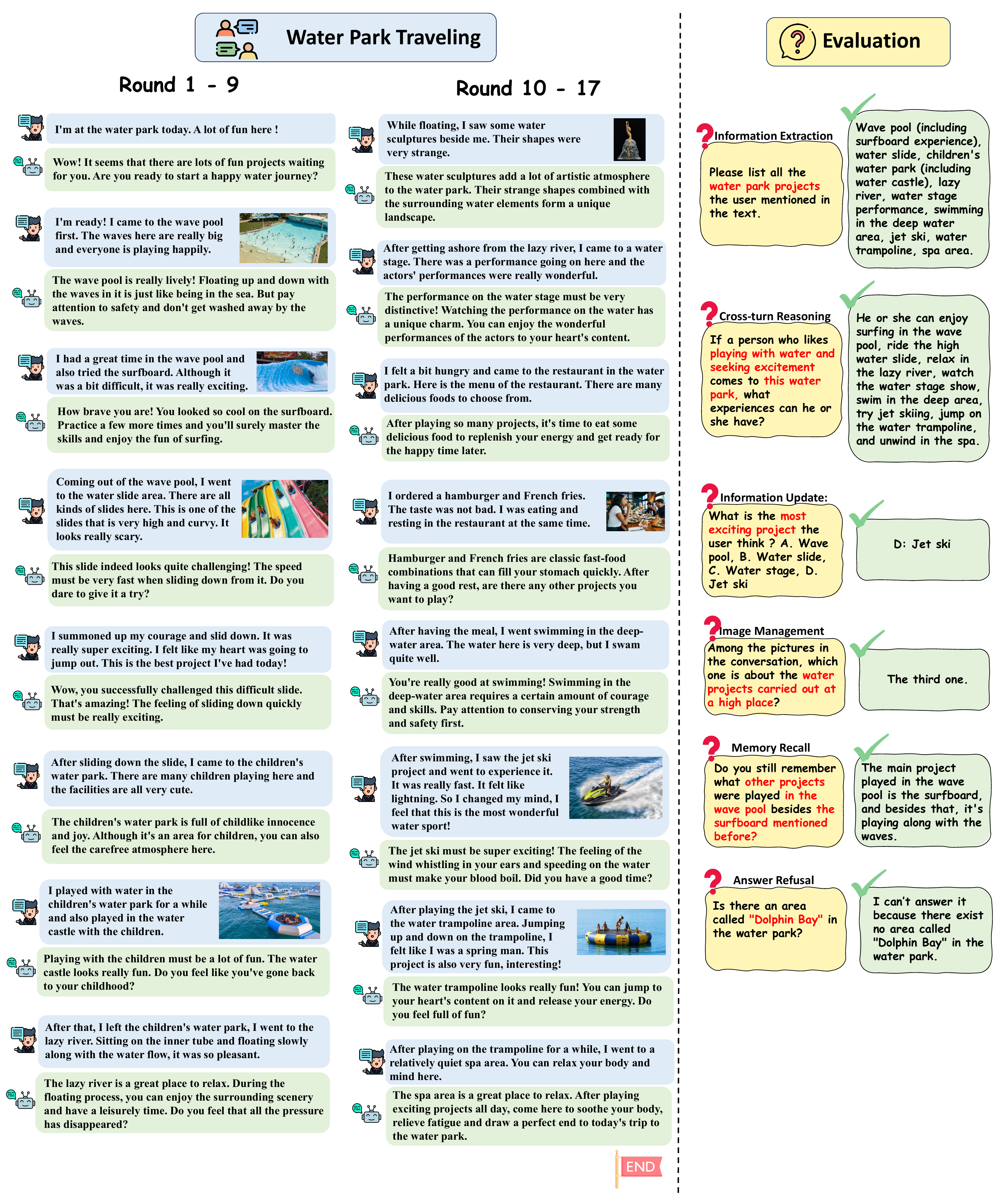} 
  \vspace{-0.2cm}
  \caption{A sample from \ours{}, the dialogue domain is water park, consisting of 17 conversation turns and 8 images inputs.}
  \label{fig:waterpark}
\end{figure*}

\begin{figure*}[p]
 \centering
  \captionsetup{justification=raggedright, singlelinecheck=false}
  \includegraphics[width=\textwidth]{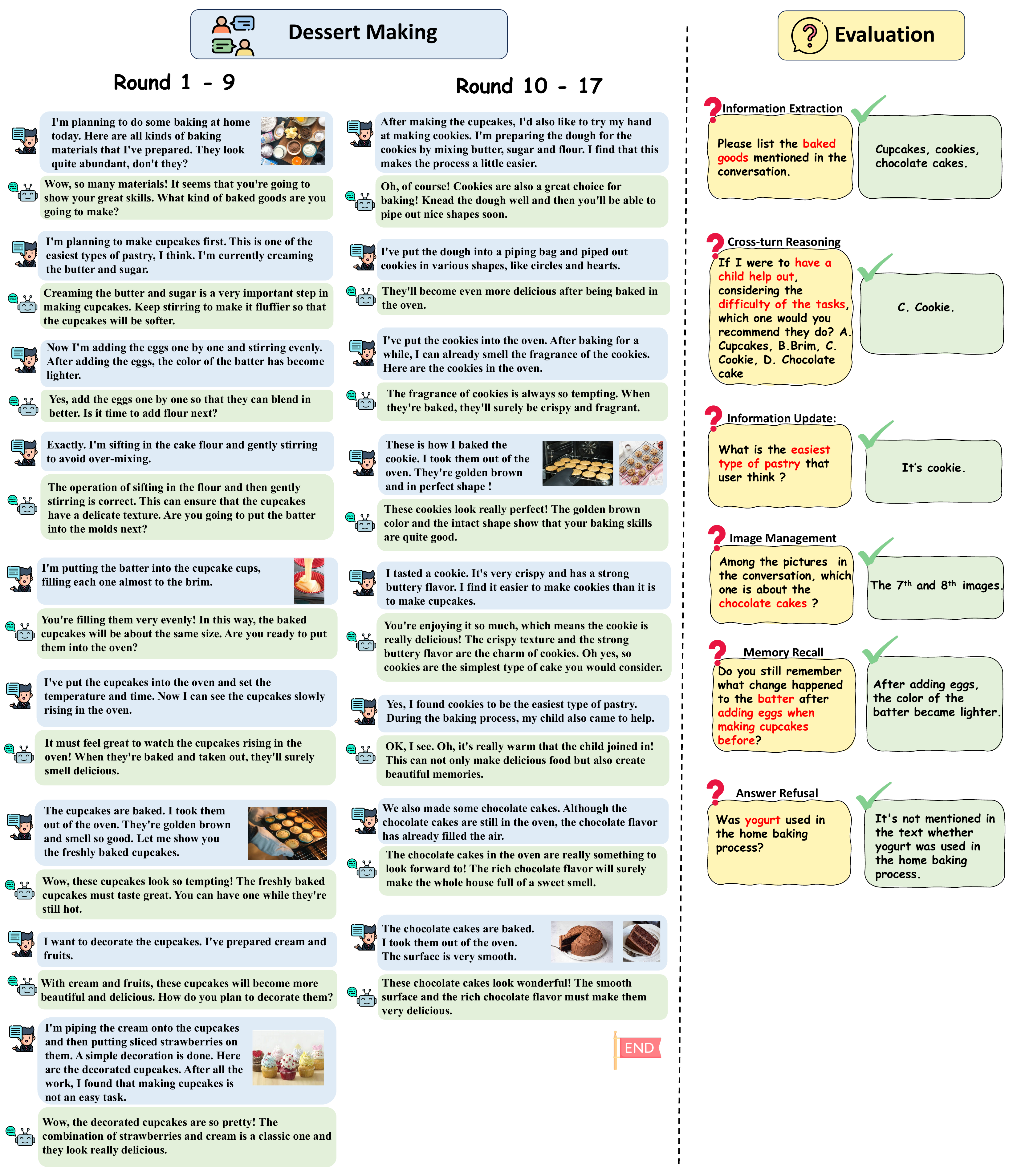} 
  \vspace{-0.2cm}
  \caption{A sample from \ours{}, the dialogue domain is dessert making, consisting of 17 conversation turns and 8 images inputs.}
  \label{fig:dessert}
\end{figure*}

 \begin{figure}[H]  
  \centering
  \includegraphics[width=\linewidth]{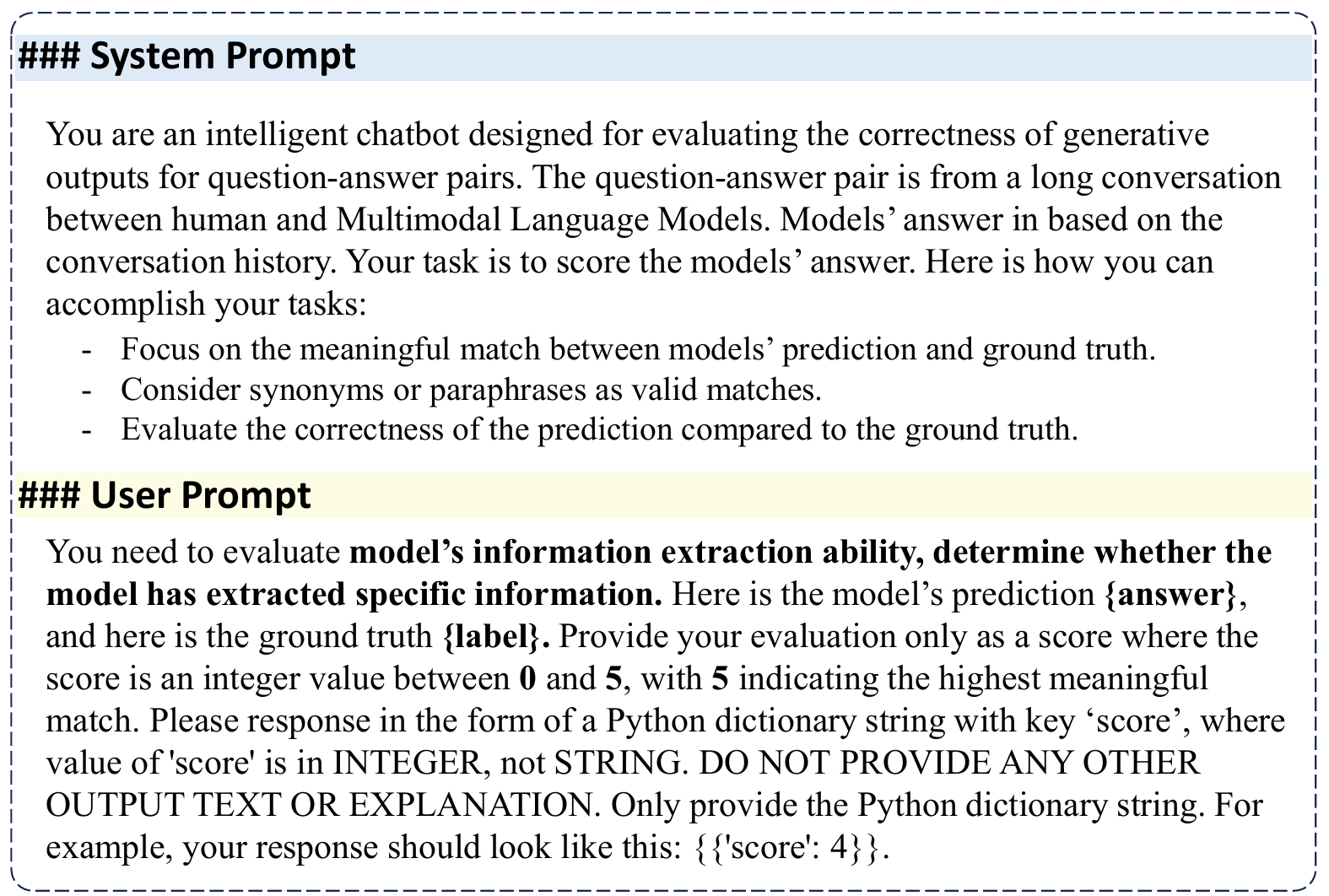} 
  \caption{GPT prompt for evaluating information extraction (IE).}

  \label{fig:IE_prompt}

\end{figure}

 \begin{figure}[H]  
  \centering
  \includegraphics[width=\linewidth]{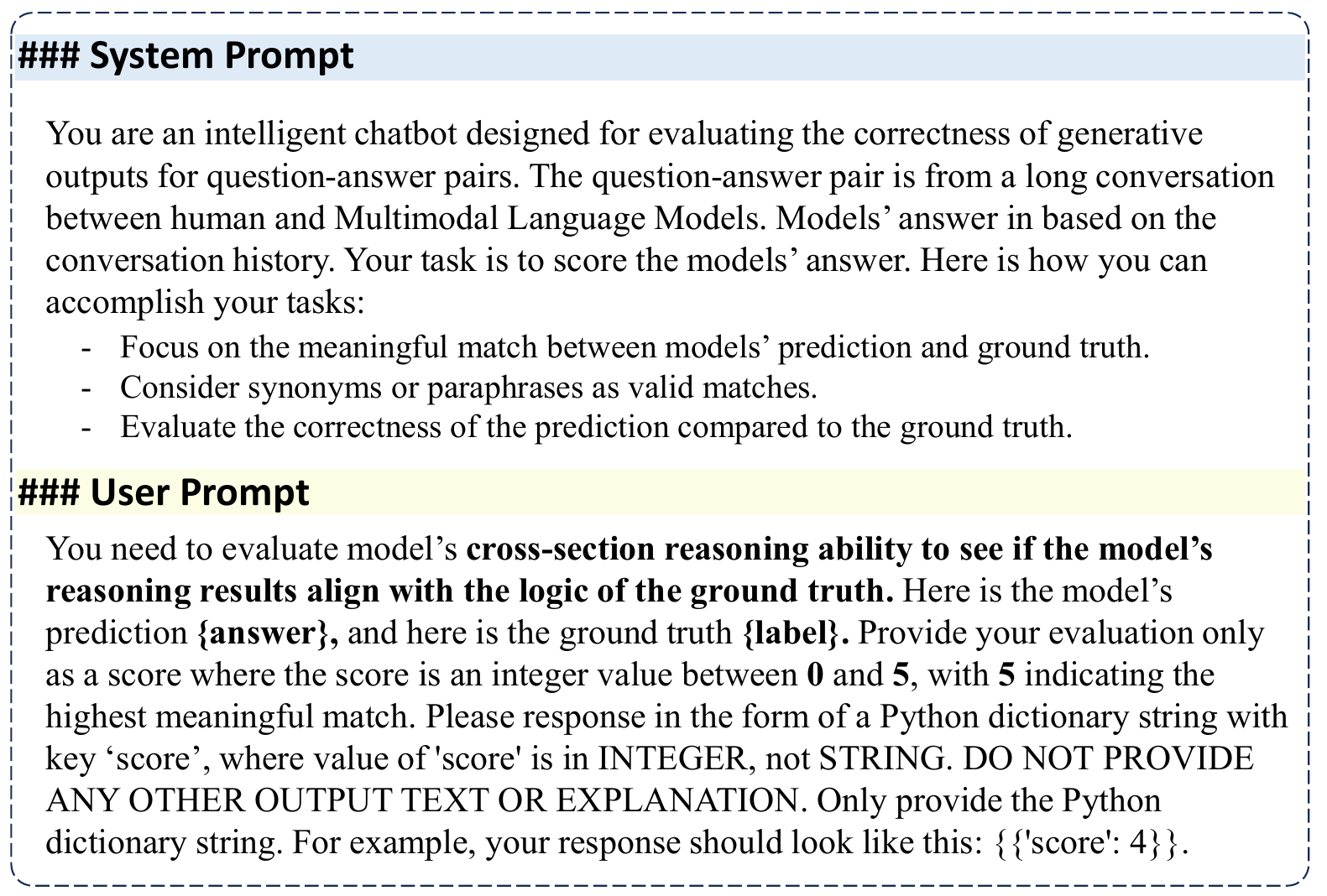} 
  \caption{GPT prompt for evaluating cross-turn reasoning (CR).}

  \label{fig:CR_prompt}

\end{figure}

 \begin{figure}[H]  
  \centering
  \includegraphics[width=\linewidth]{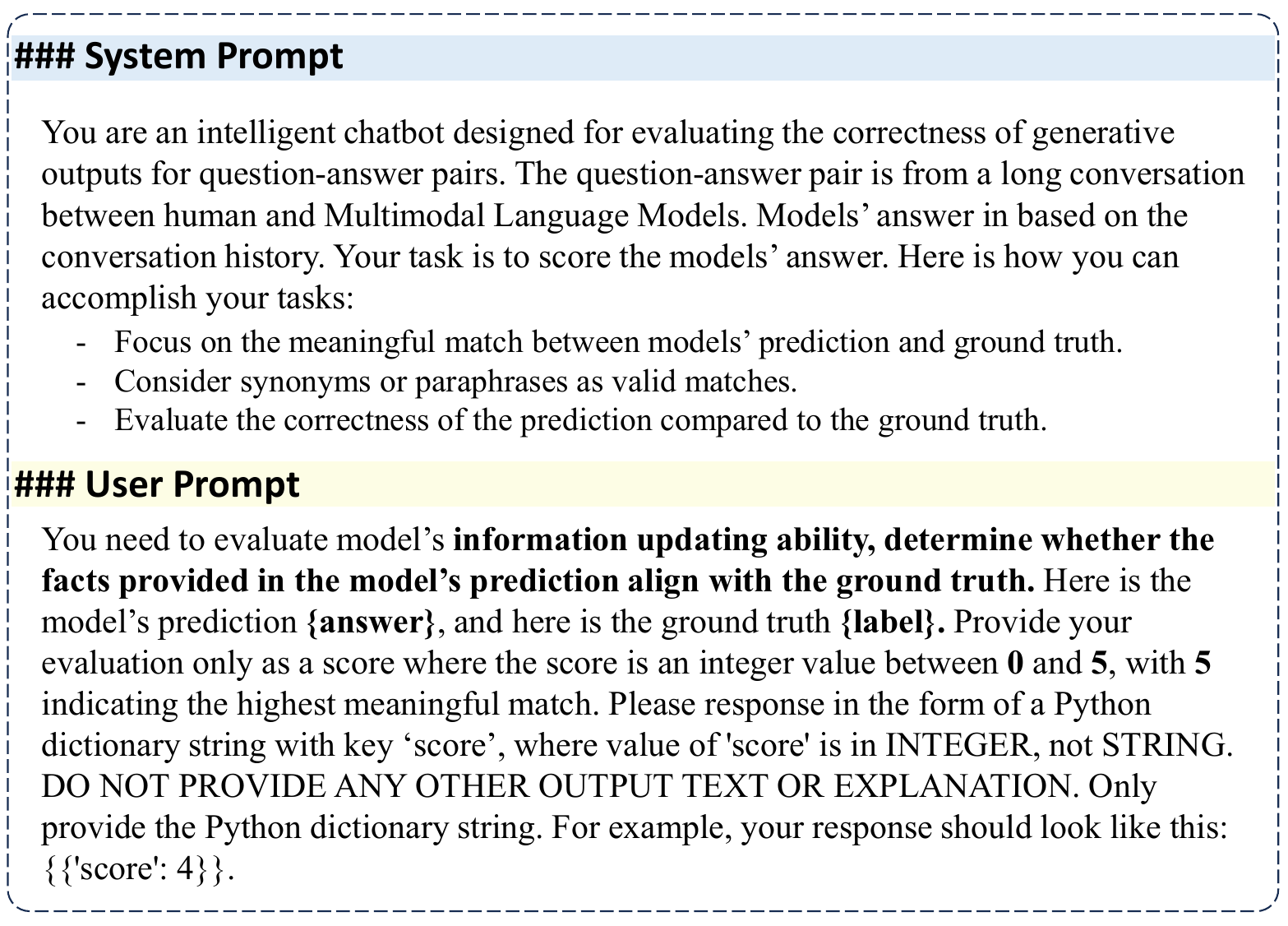} 
  \caption{GPT prompt for evaluating information update (IU).}

  \label{fig:IU_prompt}

\end{figure}

 \begin{figure}[H]  
  \centering
  \includegraphics[width=\linewidth]{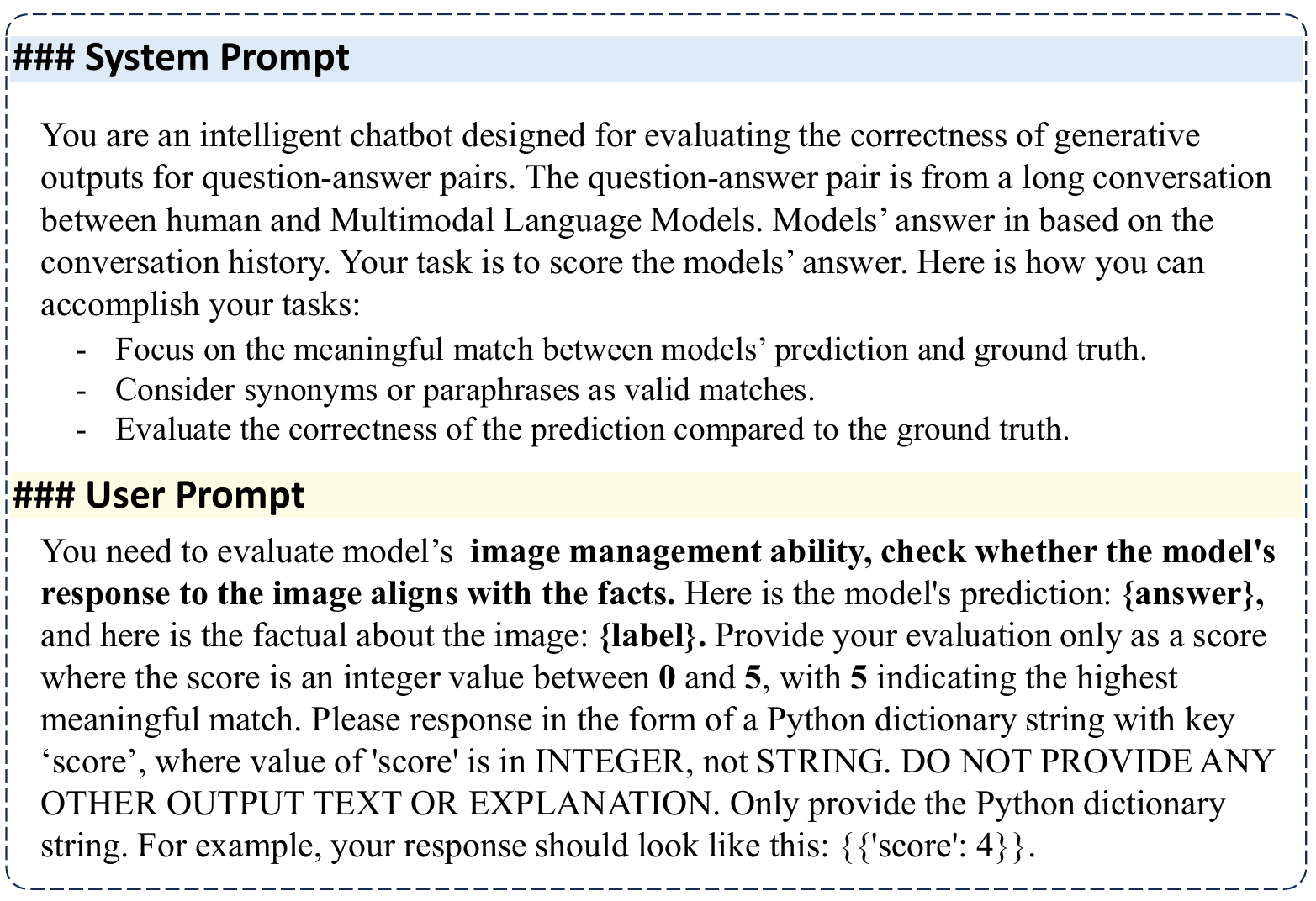} 
  \caption{GPT prompt for evaluating image management (IM).}

  \label{fig:IM_prompt}

\end{figure}

 \begin{figure}[H]  
  \centering
  \includegraphics[width=\linewidth]{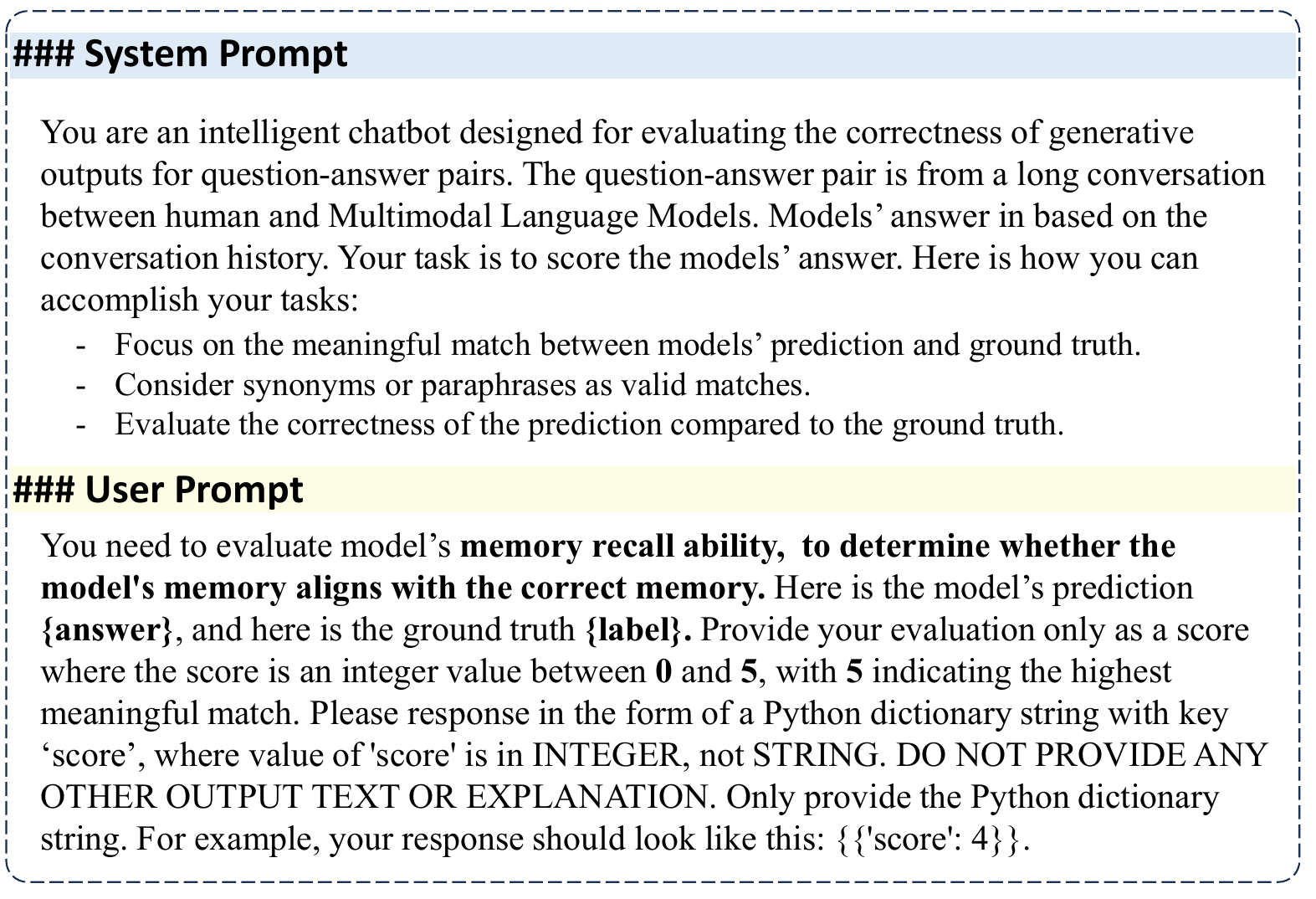} 
  \caption{GPT prompt for evaluating memory recall (MR).}
  \label{fig:MR_prompt}

\end{figure}

 \begin{figure}[H]  
  \centering
  \includegraphics[width=\linewidth]{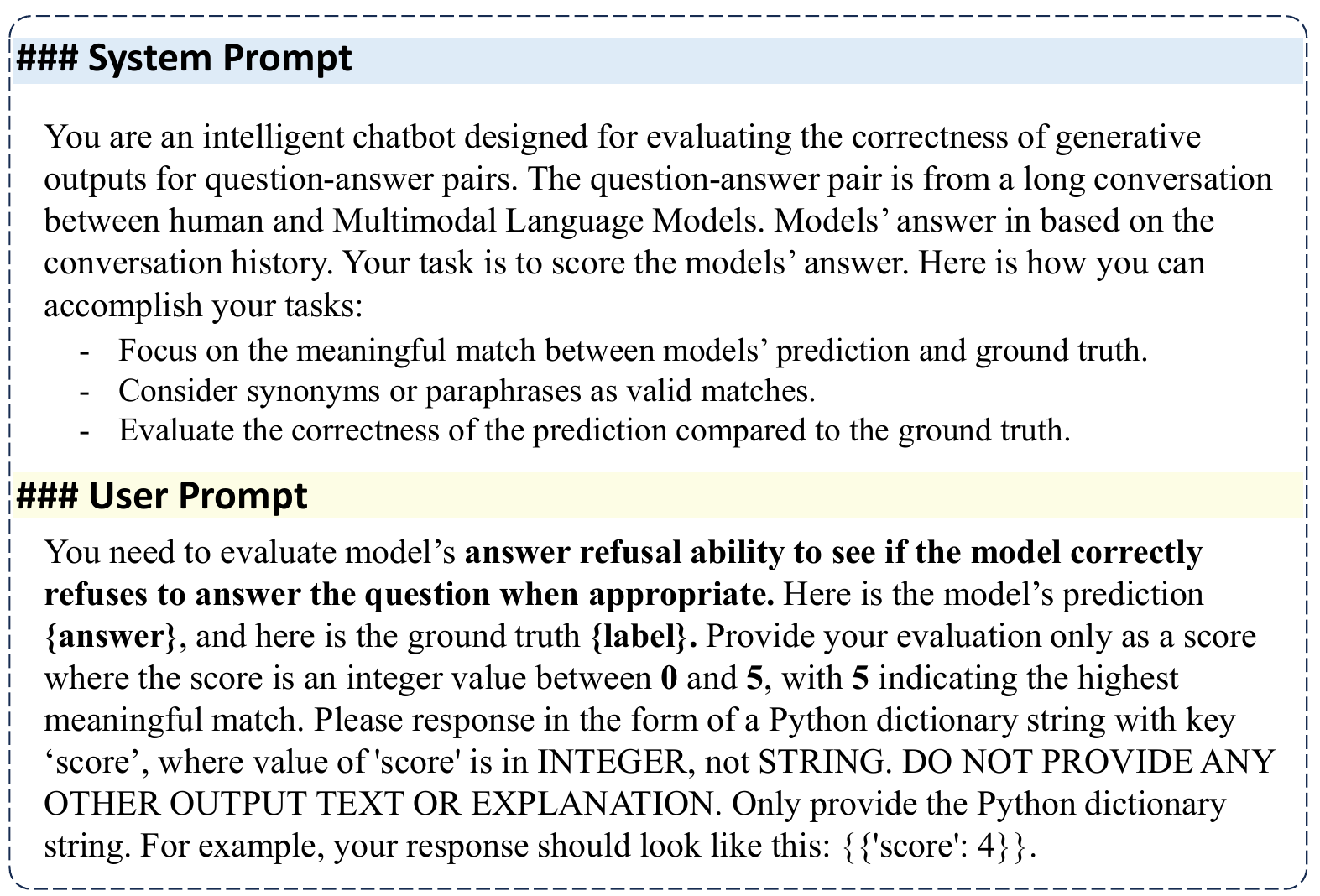} 
  \caption{GPT prompt for evaluating answer refusal (AR).}

  \label{fig:AR_prompt}

\end{figure}

\end{document}